\documentclass[10pt,journal,compsoc]{IEEEtran}
\usepackage{epsfig}
\usepackage{subfigure}
\usepackage{cuted}

\ifCLASSOPTIONcompsoc
  \usepackage[nocompress]{cite}
\else
  \usepackage{cite}
\fi
\ifCLASSINFOpdf
\else
\fi
\usepackage[cmex10]{amsmath}

\begin{document}
%
\title{Face Detection in Repeated Settings}
%
%
%
%

\author{Mohammad~Nayeem~Teli,
        Bruce~A.~Draper,
        and~J.~Ross~Beveridge
\IEEEcompsocitemizethanks{\IEEEcompsocthanksitem Mohammad Nayeem Teli is with the Department
of Electrical \& Computer Engineering, University of Maryland, College Park,
MD, USA.\protect\\
E-mail: nayeem@cs.umd.edu
\IEEEcompsocthanksitem J. Ross Beveridge and Bruce A. Draper are with Colorado State University, Fort Collins, CO, 80523, USA.}
\thanks{}}

%
%

\markboth{}%
{Shell \MakeLowercase{\textit{et al.}}: Bare Demo of IEEEtran.cls for Computer Society Journals}
\IEEEtitleabstractindextext{%
\begin{abstract}
Face detection is an important first step before face verification and recognition. In unconstrained settings it is still an open challenge because of the variation in pose, lighting, scale, background and location. However, for the purposes of verification we can have a control on background and location. Images are primarily captured in places such as the entrance to a sensitive building, in front of a door or some location where the background does not change. We present a correlation based face detection algorithm to detect faces in such settings, where we control the location, and leave lighting, pose, and scale uncontrolled. In these scenarios the results indicate that our algorithm is easy and fast to train, outperforms Viola and Jones face detection accuracy and is faster to test.
\end{abstract}

}

\maketitle

\IEEEdisplaynontitleabstractindextext

%
\IEEEpeerreviewmaketitle

\IEEEraisesectionheading{\section{Introduction}\label{sec:introduction}}

Automatic face detection is the task of determining where and how many human faces, if any, are in an image. Depending on the imaging scenario, face detection can be thought of either as a solved problem or an open challenge. At the easy end of the spectrum are mug shot photos, in which faces are shown fully frontal, at a known scale, with controlled illumination, and in front of a blank background. FERET~\cite{feret} is an example of an easy data set. For easy data sets face detection is a solved problem, with the Viola-Jones algorithm~\cite{viola-jonesJournal} being the most widely used solution. At the other end of the spectrum are totally uncontrolled scenarios. In these conditions, faces may be shown from any angle, may be partially obscured, may appear at any scale 
and under  arbitrary illumination conditions, may have motion blur, and may be set in any context. JANUS~\cite{klare2015pushing} is an example of a data set at the harder end of the spectrum, and recent studies suggest that face detection in uncontrolled scenarios is still an open research problem~\cite{cheney15unconstrained}.

\begin{figure*}[htbp!]
\centering
\subfigure[Image 1]{
\label{im1}
\includegraphics[width=2.25in]{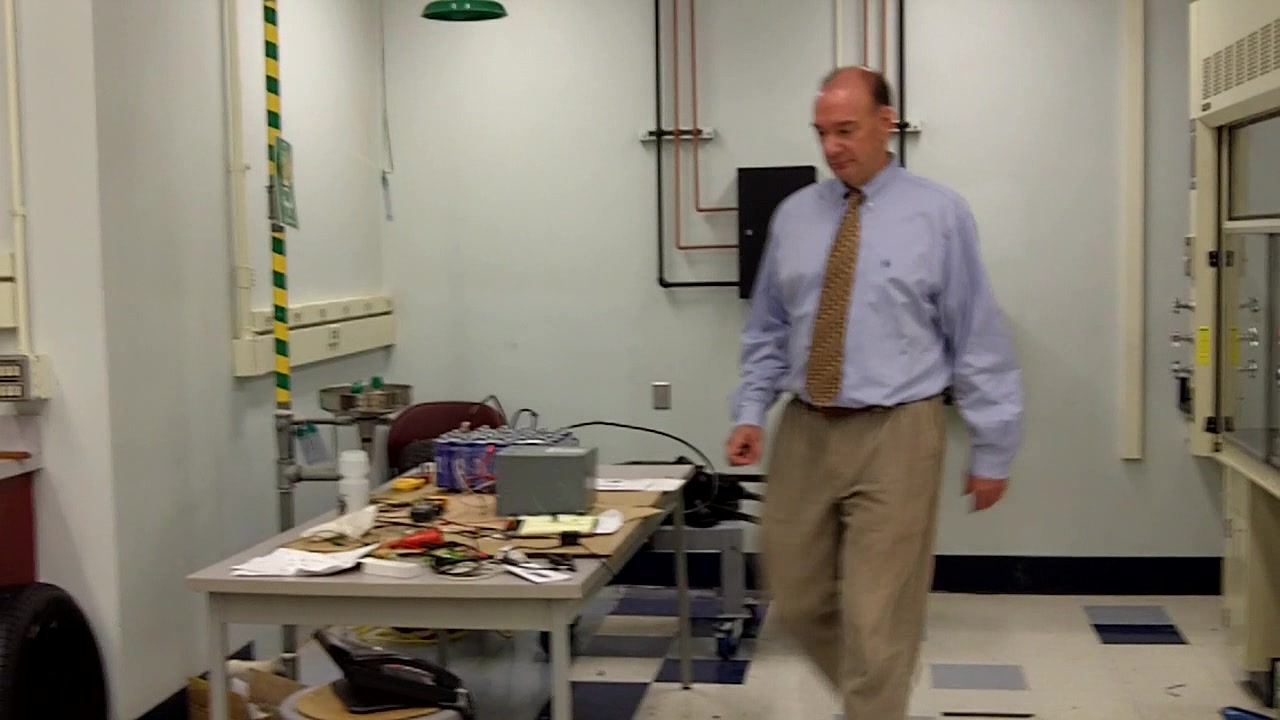}
}
\subfigure[Image 2]{
\label{im2}
\includegraphics[width=2.25in]{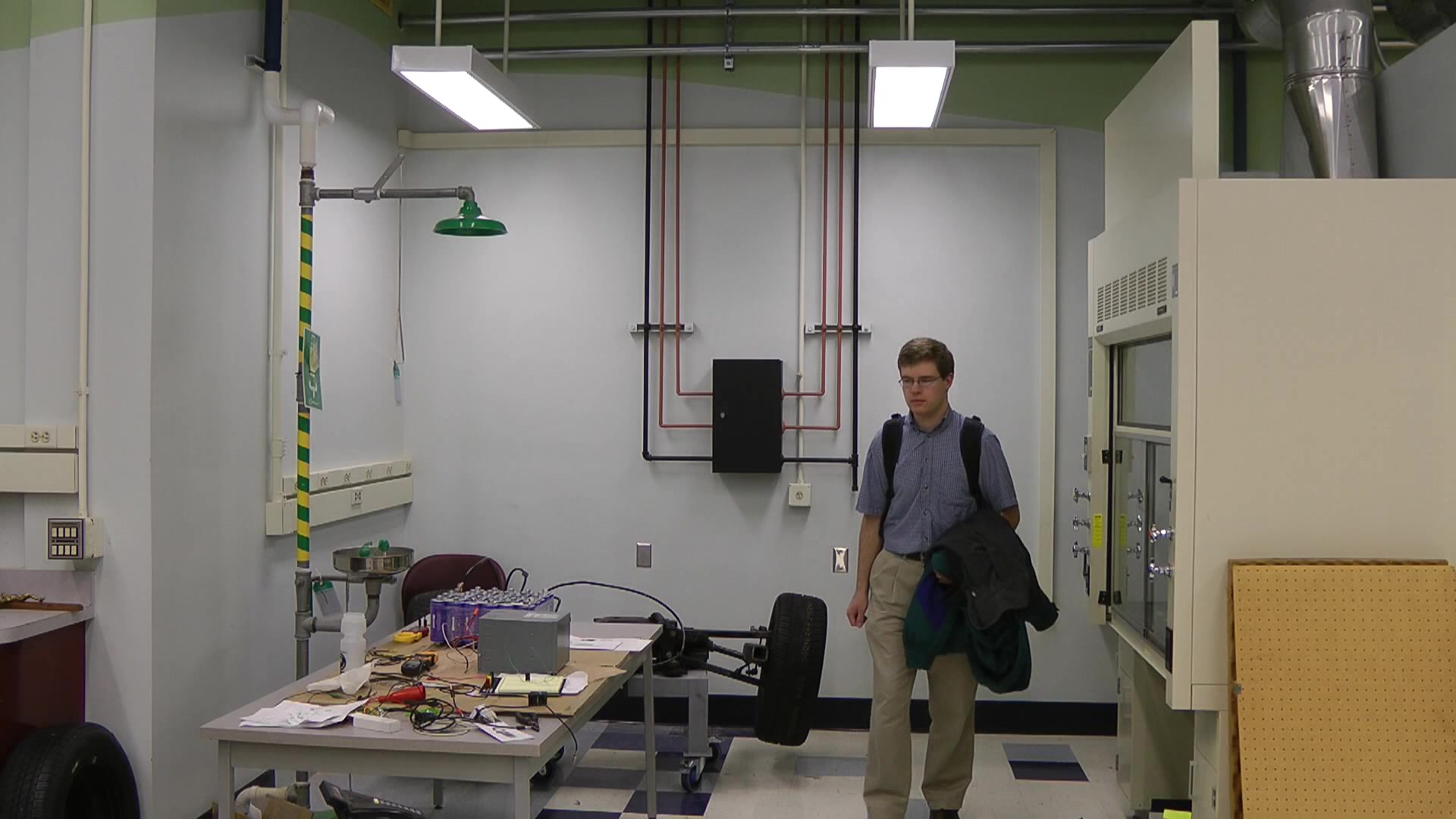}
}
\subfigure[Image 3]{
\label{im3}
\includegraphics[width=2.25in]{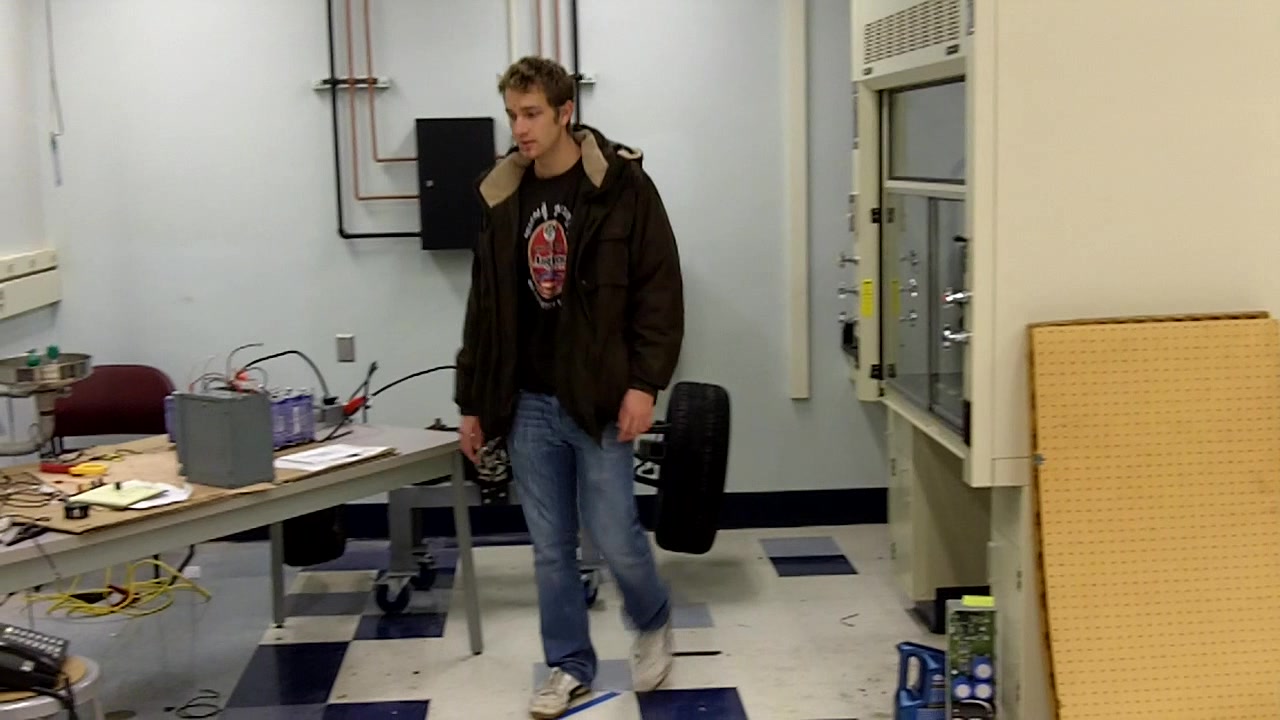}
}\\
\subfigure[Image 4]{
\label{cor1}
\includegraphics[width=2.25in]{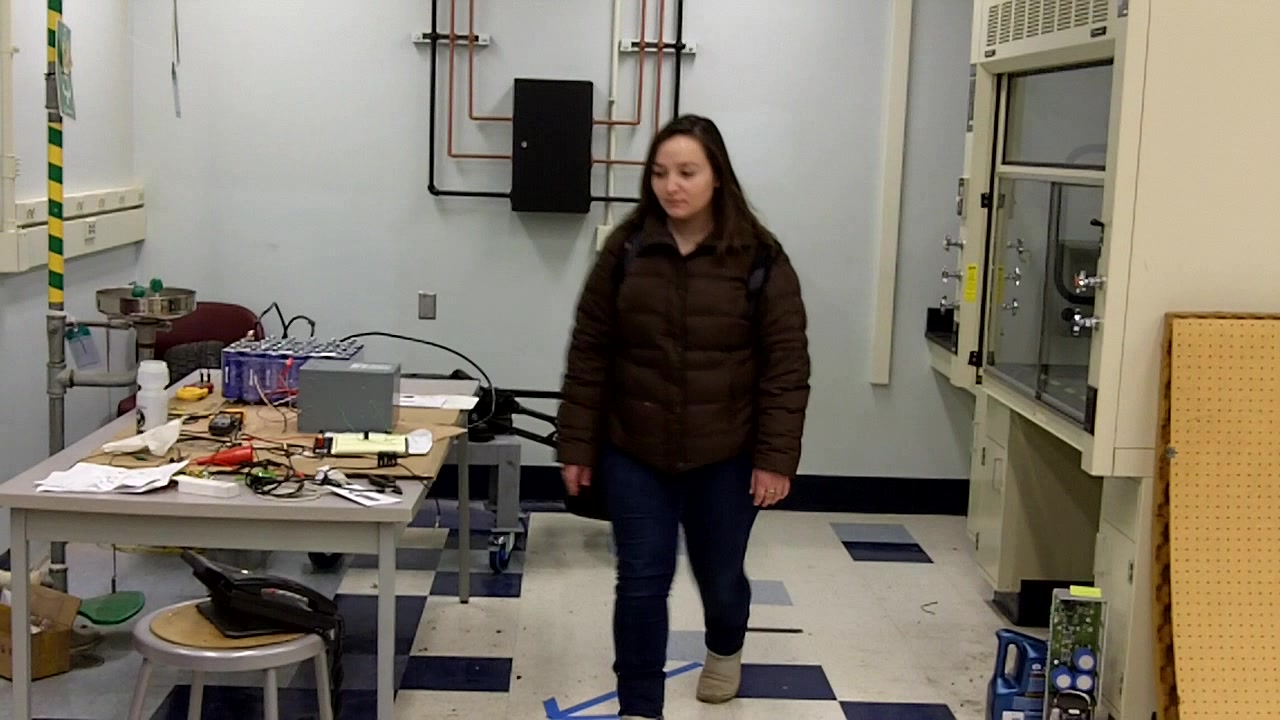}
}
\subfigure[Image 5]{
\label{cor2}
\includegraphics[width=2.25in]{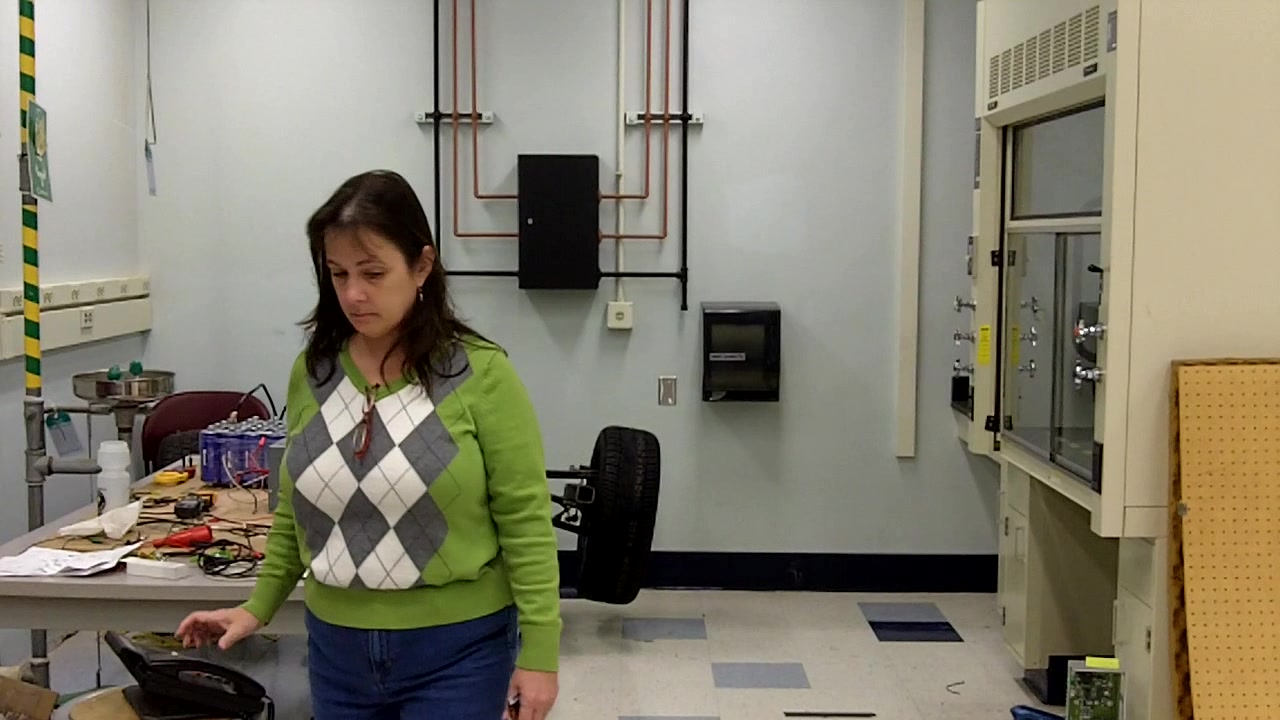}
}
\subfigure[Image 6]{
\label{cor3}
\includegraphics[width=2.25in]{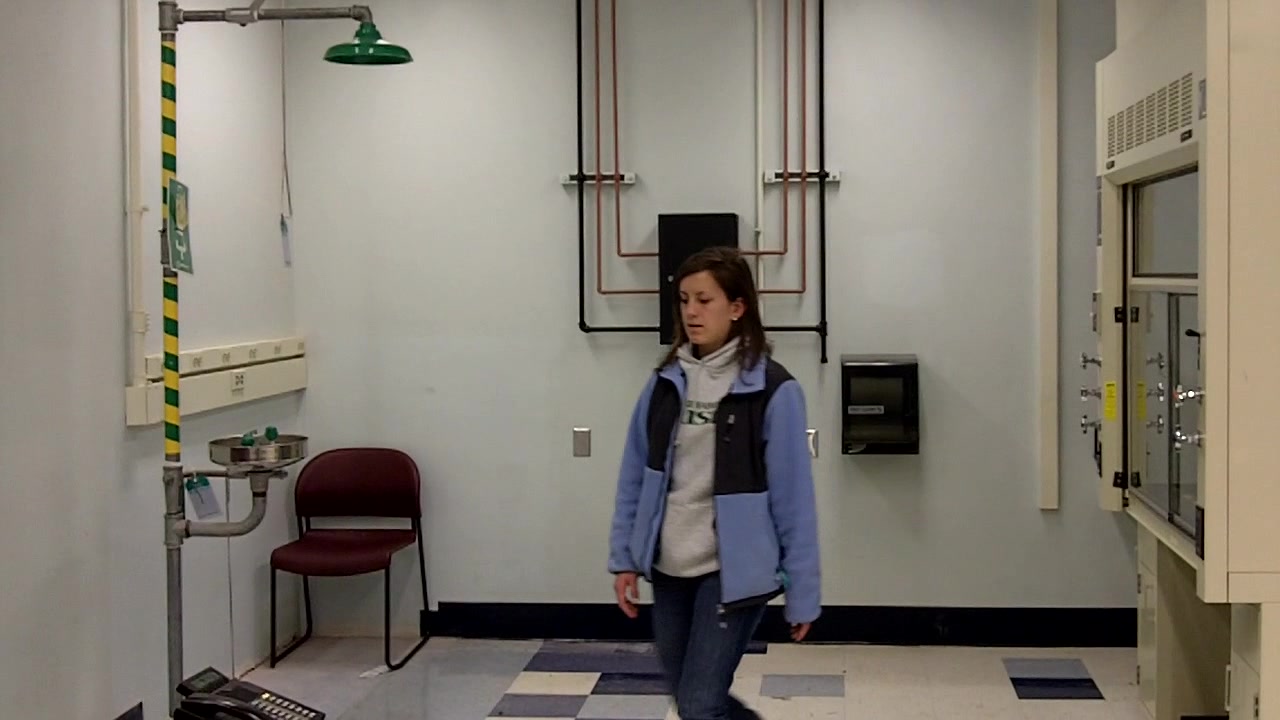}
}
\caption{Representative frames from the custom Video dataset at a particular location.}
\label{customFrames}
\end{figure*}

This paper addresses an intermediate point on the face detection difficulty spectrum. It is motivated by security applications that need to detect faces as a precondition for recognizing people. For example, an application might need to detect the faces of people in a waiting room or of people walking down a corridor. These applications lead to images like the ones shown in Figure~\ref{customFrames}. Under these conditions, faces are still seen from a wide range of angles and in motion, with variations in illumination. Occlusions also still occur. The variations in scale, however, are more limited. More importantly, although the immediate background behind the face varies as the subject walks, the overall context is constrained.

In security applications like these, the standard Viola-Jones face detector is only about 70\% accurate (see Section~\ref{ssfd} for details). Higher accuracies might be achieved by more recent techniques such as deep learning~\cite{deeplearningnature, Farfade:2015} or deformable parts models \cite{felzenszwalbcvpr10}, but they are computationally complex. Beyond recognition accuracy, one  goal for security applications is to push the computation forward toward the camera, so that only face chips (and not whole images) have to be sent over the network to main processors. This favors simple algorithms with regular computations that can be run on inexpensive, low power processors.

This paper presents a novel face detector, Correlation Face Detector (CorrFaD), based on the  Minimum Output Sum of Squared Error (MOSSE;~\cite{bolmecvpr10}) algorithm for training correlation filters. CorrFaD is a fast and simple face detection algorithm that works well in the security applications described above. Two of the factors constrained in these applications are the background and the scale. Under these constraints, we show that CorrFaD is highly reliable and we make three claims: 
\begin{enumerate}
\item CorrFaD outperforms the Viola-Jones face detector in surveillance settings such as the ones shown in Figure~\ref{customFrames}.
\item CorrFaD is fast. The time complexity to detect a face is $O(n~log (n))$.
\item CorrFaD is the first algorithm specifically tested on faces in a newly released dataset, the Point and Shoot Face Recognition Challenge (PaSC)~\cite{pasc}.

\end{enumerate}

The rest of this article is organized as follows: Section 2 introduces correlation and MOSSE correlation filter, and Section 3 presents some recent work on face detection. CorrFad and its scale sensitivity is discussed in detail in Section 4. In section 5 we present CorrFaD training and evaluation methods. Section 6 presents results of face detection in specific settings and section 7 briefly summarizes the limitations of CorrFaD.
Finally, in Section 8 we discuss our findings.
 


\section{Background: Correlation \& Filters}

In this section, we remind readers of the relevant properties of correlation and convolution. We then briefly review the state of the art in algorithms for learning optimal linear filters, with a particular focus on MOSSE~\cite{bolmecvpr10}, the algorithm at the heart of CorrFaD.

\subsection{Correlation and Convolution}
Cross-correlation
is a similarity measure between two signals when one has a time-lag. In continuous domains it is expressed as: 
\begin{equation}
(g * h)(t) \equiv  \int_{-\infty}^{\infty} g^*(\tau) h(t+\tau)d\tau,
\end{equation}
where $g$ and $h$ are continuous functions and $g^*$ is the complex conjugate of $g$. 
Convolution is similar, although one signal is reversed: 
\begin{equation}
fconv \equiv g(-t) * h(t)
\label{cc}
\end{equation}


Convolution and cross-correlation have two key features: shift
invariance and linearity. Shift invariance means that the same operation is performed at every point in the image
and linearity means that every pixel is replaced with a linear combination of its neighbors. Mathematically, linearity is represented as: 
\begin{equation}
g*kh = k(g*h)
\end{equation}
and 
\begin{equation}
g*(h1+h2) = g*h1 + g*h2.
\end{equation}

Correlation is often used to find parts of an image that match a template or {\em filter}.
Since correlation is a similarity measure, maxima in the correlation surface represent locations where the image and filter are similar. However, the limits of correlation as a matching algorithm are well known. Of importance to this paper is the bias that image patches with high values tend to produce higher correlation scores, regardless of how well they match the filter. This can be addressed by normalizing the correlation scores by the standard deviation of the image patch. 

\subsection{Discrete Fourier Transform for Correlation}
The computational cost of convolving an N$\times$N image with an M$\times$M filter is $N^2M^2$. The
cost can be reduced by taking the Discrete Fourier Transform (DFT) of both the image and
the filter using the Fast Fourier Transform (FFT) algorithm.  According to the convolution theorem, the DFT of the convolution of two functions is
the element-wise product of the DFT of one function with the complex conjugate of the DFT of the other,
as shown below:
 \begin{equation}
Corr(g,h) \leftrightarrow G(f)H^*(f)
\label{eq:frequency_correlation}
\end{equation}
where $G(f)$ and $H^*(f)$ are the Discrete Fourier Transform of $g(t)$ and the complex conjugate of the Discrete Fourier Transform of $h(t)$, respectively. 
The computational complexity of converting images into the frequency domain, computing their correlation as in Equation~\ref{eq:frequency_correlation}, and converting them back to the spatial domain (if necessary) is only $O(n\hspace{0.05cm} log \hspace{0.05cm}n)$. This makes correlation a computationally efficient, if not necessarily accurate, matching algorithm. 

\subsection{Correlation Filters}
\label{sec:corrfilters}
Because of their speed, correlation filters~\cite{corrFil} continue to be used in applications such as face verification~\cite{savvides}, face recognition~\cite{kumar2007}, and target detection~\cite{goo96}. Recent research has focused on improving performance by learning filters that will give the best possible results. 

Matched filters~\cite{north63} convolve an unknown signal with the conjugate of a time reversed known template to determine whether the template exists in the unknown signal. The goal is to maximize the Signal-to-Noise (SNR) ratio in the presence of additive stochastic noise.

Although matched filters are usually used in RADAR applications, they also find use in improving SNR in X-rays. 
Their use is limited in face detection applications because they require a large number of filters to account for variations in expressions, pose, age, illumination etc.,~\cite{kumar2007}.

Synthetic Discriminant Function (SDF) filters~\cite{sdf} are used for face verification. They are trained to produce correlation values of one for positive training samples and zero for negative training samples. A drawback of SDF filters is that they require carefully centered and cropped training samples in order to compute the optimal filter values~\cite{kumar2007}.

Minimum-Variance Synthetic Discriminant Functions (MVSDF) filters~\cite{mvsdf} satisfy similar constraints to SDF filters~\cite{sdf} while minimizing the output variance due to white noise by removing the condition that the filter must be a linear combination of the positive training samples.  

Minimum Average Correlation Energy (MACE) filters~\cite{mahalanobis} minimize the average energy of the correlation surface over the training images. It is designed such that the correlation value is highest at object locations  and zero everywhere else. MACE filters are sensitive to input noise and deviations from the training images because they emphasize high spatial frequencies in order to create sharp correlation peaks~\cite{kumar2007}. 
MACE filters have been successfully applied to face recognition and face verification~\cite{savvides03,kumar02,savvides,savvidesKumar}.

Optimal trade-off filters (OTF)~\cite{refregier} optimize a combination of three performance measures: SNR to measure noise tolerance, peak-to-correlation energy to measure peak sharpness, and Horner efficiency~\cite{horner}~\cite{caulfield} to measure the light-throughput efficiency of the filter. OTF filters can be set to optimize any one of these three measures while holding the other two at specific values.

Unconstrained Minimum Average Correlation Energy (UMACE) filters~\cite{umace} create sharp correlation peaks for easy output detection even in the presence of noise and background clutter. Unlike other correlation filters, UMACE filters are not constrained to produce a pre-specified correlation value at a specific point.

Still more correlation filters can be found in the literature. Distance Classifier Correlation Filters (DCCF)~\cite{dccf} are designed to separate training images from different classes into well-separated clusters.
Mahalanobis and Kumar~\cite{polyfilter} present polynomial correlation filters that convolve a filter with non-linear functions of the input image.  Recently, Kerekes and Kumar~\cite{kerekes} extended the correlation filter design using Mellin radial harmonic functions~\cite{mellin}. In this research we use MOSSE which is described next.


\subsection{MOSSE Filters}

With minor variations, the correlation filters described above are
trained on positive and negative samples, where samples are filter-sized image windows that either are (positive) or are not (negative) centered on the target of interest. In our case, the targets of interest are faces. Typical training images, however, contain only a few image windows that are centered on faces but a large number of image windows that aren't. Recently, new algorithms for training filters have been developed that take advantage of all of these negative samples and therefore perform particularly well in the context of known backgrounds. Historically, the first of these filters was ASEF~\cite{asef}, followed by MOSSE~\cite{bolmecvpr10}. Since MOSSE filers are the foundation of CorrFaD, we take time to review MOSSE in detail here. 

\begin{figure*}[htbp!]
\includegraphics[width=7.0in]{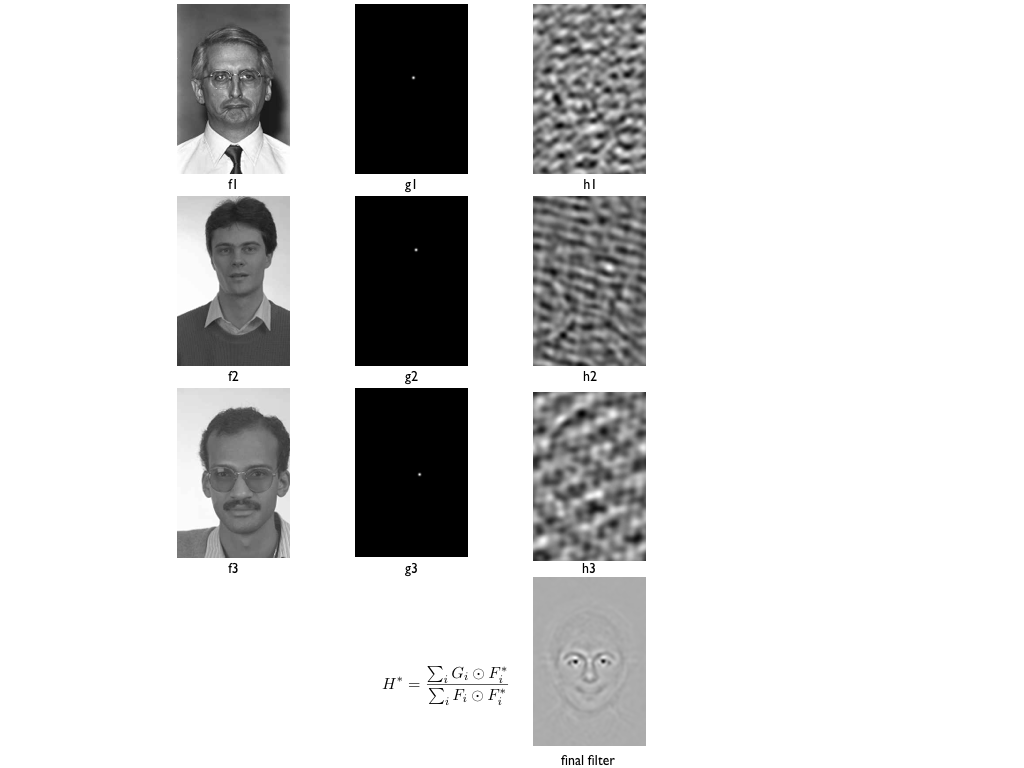}
\caption{MOSSE Filters.  The left column shows three very simple face images. The middle column shows the idealized outputs given these images, while the right column shows the corresponding (and overfit) exact filters. The bottom right image shows the MOSSE filter created from the data set using Equation~\ref{finalmos}.}
\label{mossedesign}
\end{figure*}

\label{mosse}
Unlike the filters described in Section~\ref{sec:corrfilters}, 
MOSSE~\cite{bolmecvpr10} filters are trained on whole images that may contain zero, one or more faces\footnote{MOSSE filters can be trained to recognize other objects than faces, but faces are the targets of interest in this paper.}. Instead of extracting sample windows from the training images, the user instead supplies idealized output images. Since the goal of a filter is to only respond to faces, the idealized output images are zero almost everywhere. However, where faces occur in the source image, the idealized output has a Gaussian impulse centered at the center of the face. Figure~\ref{mossedesign} shows three simple source images and the corresponding idealized output images.

The goal is to learn a filter $h$ that when correlated with training image $f$ will produce the given output image $g$. The
filter is designed in the Fourier domain to exploit the convolution theorem. In the Fourier domain, the exact filter for a given training image and output image is $H$:
\begin{equation}
G = F  \odot H^*
\end{equation}
where $\odot$ denotes element-wise multiplication and $*$ is the complex  conjugate.  This can be easily solved for the exact filter by:
\begin{equation}
H_i^* = \frac{G_i}{F_i}
\end{equation}

The right column of Figure~\ref{mossedesign} shows the exact filters for three simple face images. The filters are non-intuitive because they are overly specific. They contain lots of frequencies whose sole purpose is to exactly cancel out the background, and for a single image there is always a filter that will exactly recreate the idealized output image. The goal, however, is to learn a single filter $H$ that will come close to recreating the idealized output image corresponding to every training image. Thus the goal is to minimize
\begin{equation}
min_{H^*}\sum_i|F_i \odot H^* - G_i|^2,
\label{sse}
\end{equation}
where every training image has a unique value $i$. It is shown in \cite{bolmecvpr10} that this equation is minimized by 
\begin{equation}
H^* = \frac{\sum_iG_i \odot F_i^*}{\sum_iF_i \odot F_i^*}
\label{finalmos}
\end{equation}
The bottom right image in Figure~\ref{mossedesign} shows the resulting and more intuitive filter.

Before computing the MOSSE filter as above, the training images are pre-processed as in~\cite{asef}. First, pixel values are transformed using the
natural log function to help with low contrast lighting situations. Second, pixels are normalized  such that the image has a
mean of zero and a magnitude of one. Third, the image is multiplied by a cosine window. Such a window gradually reduces the pixel values
near the edges to zero, putting more emphasis on the center of the image. 

Having presented the correlation filter underlying our face detector, we next give a brief review of face detection research before introducing CorrFaD.   

\section{Background: Face Detection}
The first step in face processing is to detect faces. Face detection determines whether an image contains faces, and if it
does, reports the locations of each face~\cite{yangkriegmanahuja}. Face detection is an active area of research including exemplar-based face detection~\cite{Lincvpr2014} and cascade-based face detection~\cite{chencvpr2014}. Recently, convolutional deep neural networks have been used for face detection~\cite{Licvpr2015} \cite{yangiccv2015} \cite{Farfade:2015} including a multi-tasked cascaded convolutional network~\cite{zhang2016}.

The recent work in face detection, including most of the citations above, has concentrated on finding faces under more and more challenging conditions. In terms of the introduction to this paper, these papers have concentrated on the hard end of the face detection spectrum. This is understandable, since face detection in mug shots and other easy settings is a solved problem. To address ever more challenging scenarios, however, these works have adopted techniques such as deep neural networks that consume significant computational resources. 

We believe that many security applications, however, confront scenarios with multiple cameras but limited network bandwidth. In these cases, it is desirable to use computationally simple face detection algorithms that can be implemented near the camera, so that only detected faces have to be forwarded over the network. At the same time, the face detection algorithms have to be sophisticated enough to detect faces as they wander around a room or walk down a corridor. These security applications are not fully on the easy side of the face detection spectrum, either.



The most common inexpensive face detection algorithm is the Viola and Jones~\cite{viola-jones}~\cite{viola-jonesJournal} face detector. This algorithm can detect faces at 15 frames per second on traditional hardware. Special-purpose hardware can make it faster, and because of its speed it has been embedded in many point-and-shoot cameras. 
The Viola-Jones algorithm is notable for its integral image representation~\cite{crow1984}, the use of AdaBoost~\cite{freund} for training, and its cascade of classifiers. The integral images and the classifier cascade, in particular, account for its speed. 

This paper presents a new algorithm based on correlation that, like Viola-Jones, is fast enough and simple enough to embed in a camera. Unlike Viola-Jones, however, it can detect faces when people are moving around inside an environment like a waiting room where the pose and illumination change but there are background patterns that can be learned and discounted. In essence, CorrFaD learns correlation filters that respond to faces but not to the environment. As a result, it is able to outperform Viola-Jones without requiring significantly more computational resources.

\section{Correlation-based Face Detection}
Applications on the easy end of the face detection spectrum are well-served by fast algorithms, most notably Viola-Jones~\cite{viola-jonesJournal}. Our goal in developing CorrFaD was to push fast face detection techniques more toward the middle of the spectrum, and in particular to create a fast technique to detect faces in complex but repeated settings, for example waiting rooms and corridors. 

Our hypothesis is that better algorithms for training optimal filters (as discussed in Section~\ref{sec:corrfilters}) enable face detection algorithms based on image correlation. Image correlation (a.k.a. template matching) is, of course, the oldest technique in computer vision. The process is well known. One simply convolves an image with a filter and thresholds the local maxima in the resulting convolution image. If the image size (i.e. scale) of the target is unknown, one can use image pyramids~\cite{burtAdelson} to rescale the source image and efficiently locate targets at multiple scales.

The limits of image correlation are also well known, of course. One problem is setting the appropriate threshold. In most domains, the appearance of a target will change from image to image. In face recognition, for example, the appearance of a target is largely a function of viewpoint. We also expect illumination changes as a person walks around a room. In addition, the target object itself may change over time, as when people change expressions or hair styles. All of these factors degrade the response of the filter to faces, and it can be difficult or even impossible to find a threshold that separates the responses of true faces from the random responses of background objects.

MOSSE filters, however, minimize the problem of false responses to background objects. 
The MOSSE filter training algorithm explicitly creates filters that respond to faces but not to the background. While all the correlation filter training algorithms discussed above try to maximize target responses relative to background responses, none make comprehensive use of the backgrounds of the training images the way MOSSE does. As a result, as long as the training images are acquired in the same setting as the test images -- for example, the same waiting room or corridor --  nothing in the background should respond strongly to the MOSSE filter.

Geometric transformations remain the other major problem. Correlation is insensitive to changes in translation, but is unfortunately sensitive to changes in rotation or scale. Worse still, faces are non-planar objects in 3D; their 2D image appearance therefore depends on the orientation of the face in 3 space relative to the camera. 

Fortunately, in most applications we do not expect to encounter faces at truly arbitrary orientations. People tend to stand or sit mostly upright; headstands are less common. Although we often look up or down, we do this mostly by moving our eyes. The tilt of our head is relatively subdued. We do expect rotations in the horizontal plane; faces may be oriented toward the camera, but they may also be facing to the left or right. We also expect changes in scale as people move closer to or farther from the camera.

Our approach to compensating for changes in orientation and scale is simple and brute force: more filters, combined with image pyramids. The rest of this section describes experiments to test the range of scales and orientations a MOSSE face filter responds to. As will be shown, 3 filters covers an octave of scale space, and at each of these scales we need filters for three orientations. As a result, 9 image convolutions per octave is enough to detect faces in many applications.

\subsection{Scale Sensitivity}
To measure the sensitivity of MOSSE face filters to changes in scale, we first conducted a set of experiments on images from the FERET~\cite{feret} face data set. As mentioned in the introduction, FERET is an easy data set. Nonetheless, we use FERET in order to isolate the effects of scale from other factors. When we introduce the full CorrFaD algorithm in Section~\ref{corrfad}, we will evaluate its performance on more challenging data sets.

\begin{figure*}[htbp!]
\centering
\subfigure[Image 1]{
\label{im1}
\includegraphics[width=1in]{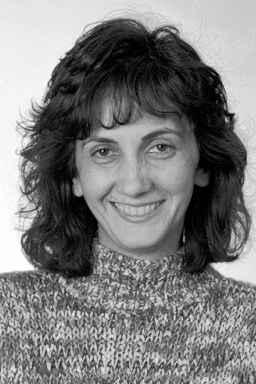}
}
\subfigure[Image 2]{
\label{im2}
\includegraphics[width=1in]{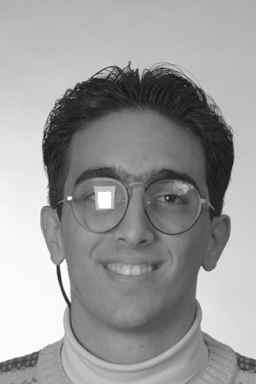}
}
\subfigure[Image 3]{
\label{im3}
\includegraphics[width=1in]{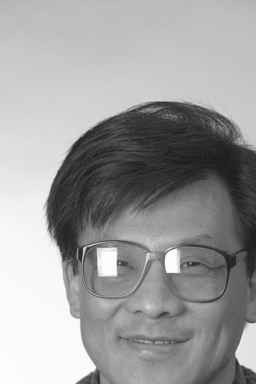}
}
\subfigure[Image 4]{
\label{cor1}
\includegraphics[width=1in]{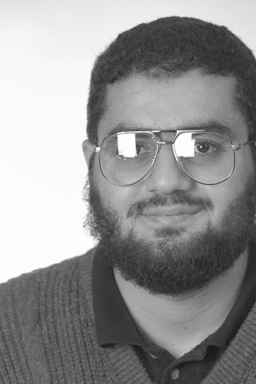}
}
\subfigure[Image 5]{
\label{cor2}
\includegraphics[width=1in]{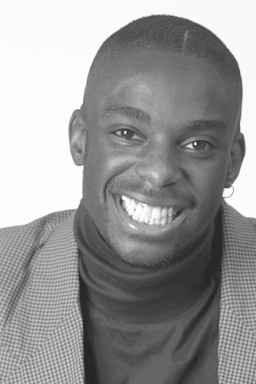}
}
\subfigure[Image 6]{
\label{cor3}
\includegraphics[width=1in]{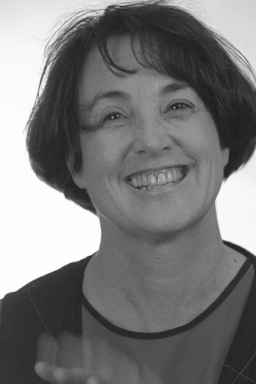}
}
\caption{Representative images from the FERET database.}
\label{feretImages}
\end{figure*}

We use 500 images selected from FERET to measure scale sensitivity. Figure~\ref{feretImages} shows 6 of the 500 images. As shown in Figure~\ref{feretImages}, FERET images have uniform backgrounds, so false matches from background features are not an issue. FERET images are also labeled in terms of pose; we use 500 frontal images so that there are no significant changes in orientation. Other sources of variation, such as changes in expression and illumination, remain. The 500 images are divided into training and test sets such that no subject appears in both the training and test set.

\subsubsection{Baseline Scale Performance}
The first experiment establishes a baseline: how well do MOSSE face filters perform when the scales of the faces in the test set, measured in terms of pixels between the eyes, are the same as the scales of the training images? To determine this, we re-scaled the images in both the training and test sets so as to make the interocular distances the same. We did this 6 times, creating data sets with interocular distances of 16, 32, 48, 64, 80 and 96 pixels. We then correlated test images with an interocular distance of 16 with a MOSSE filter trained on faces with 16 pixels between the eyes. We also correlated images with interocular distances of 32 with a filter training on faces with 32 pixels between the eyes, and so on. At each scale, we had 250 correlation tests, one for each test image. After each correlation we located the peak filter response. A trial was said to be a success if the peak was within 5 pixels of its ground truth location (the midpoint between the eyes). 
\begin{figure*}[htbp]
\centering
\subfigure[16]{
\label{16 Pixels Between Eyes}
\includegraphics[width=0.9in]{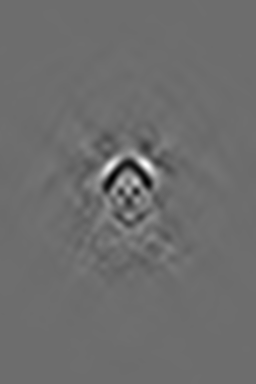}
}
\subfigure[32]{
\label{32 Pixels Between Eyes}
\includegraphics[width=0.9in]{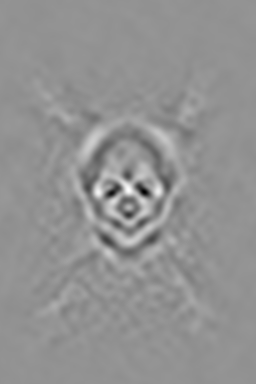}
}
\subfigure[48]{
\label{48 Pixels Between Eyes}
\includegraphics[width=0.9in]{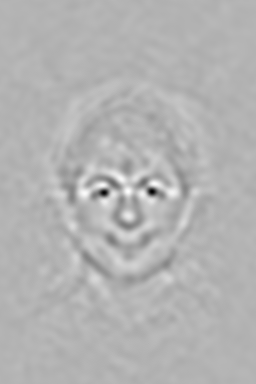}
}
\subfigure[64]{
\label{64 Pixels Between Eyes}
\includegraphics[width=0.9in]{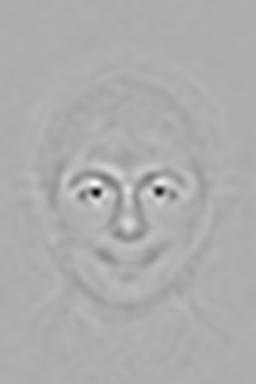}
}
\subfigure[80]{
\label{80 Pixels Between Eyes}
\includegraphics[width=0.9in]{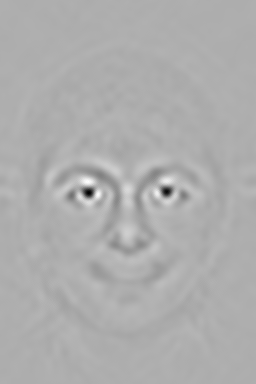}
}
\subfigure[96]{
\label{96 Pixels Between Eyes}
\includegraphics[width=0.9in]{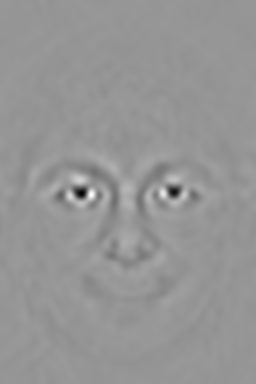}
}
\subfigure[Unprocessed]{
\label{Unprocessed}
\includegraphics[width=0.9in]{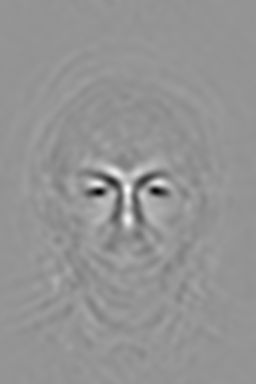}
}
\caption{Each filter is labeled as the number of pixels between the eyes in the training dataset .}
\label{filters}
\end{figure*}
\begin{table}[htp]
\caption{Accuracy of locating point between the eyes with filters trained  and tested on the images with the same number of pixels between the eyes.}
\begin{center}
\begin{tabular}{|c|c|}
\hline
Pixels between Eyes &  \% Accuracy (within 5 pixels)\\
\hline
16 &  91.2 \\
\hline
32 & 77.6\\ 
\hline
48 & 87.6 \\
\hline
64 & 95.2 \\
\hline
80 &  96.4 \\
\hline
96 & 97.2 \\
\hline
\end{tabular}
\end{center}
\label{samepixels}
\end{table}%

The results of these tests are presented in Table~\ref{samepixels}. Not surprisingly, larger faces have more information and are therefore easier to detect than smaller ones. MOSSE filters correctly localize (to within 5 pixels) 91.2\% of all faces when the interocular distance is 16 pixels; this rises to 97.2\% when the interocular distance is 96 pixels. In general, however, MOSSE filters localize faces over 90\% of the time in easy images from the FERET data set when there are no differences in scale or orientation. (There is a small, unexplained exception for faces with 32 pixels between the eyes.)  Figure~\ref{filters} shows visually what the MOSSE filters look like for different scales. The increase in detail explains the 6\% rise in accuracy over the range of interocular distances.


\subsubsection{Performance as a Function of Scale Change}
\label{sec:feretscale}
Having established a baseline, our next goal is to measure the sensitivity of MOSSE face filters to changes in scale. Invoking scale space theory, we measure scale changes in terms of octaves. Every octave represents a doubling of the pixels between the eyes. Anticipating the use of image pyramids, our goal is to determine how many filters are needed per octave.

\begin{figure}[htbp]
\begin{center}
\includegraphics[width=2.5in]{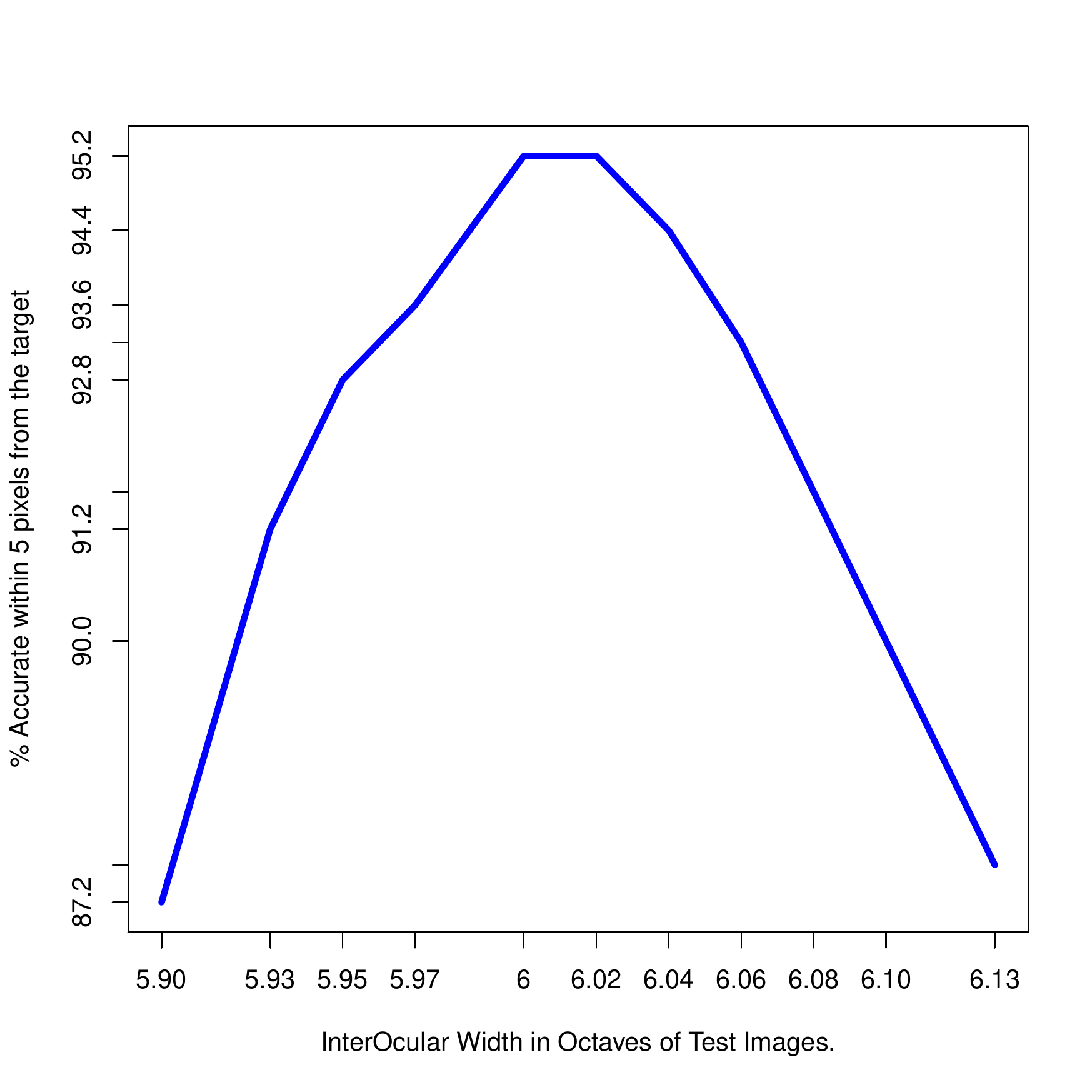}
\caption{Accuracy of locating the target within 5 pixels when a filter is trained on 6 octave and tested on 13 datasets containing images between octaves 5.85 and 6.13. }
\label{octave6plot}
\end{center}
\end{figure}

We begin by re-scaling the training images to an interocular width of 64 pixels (6 octaves) and training a MOSSE filter for faces as that scale. We then re-scale the test images to different interocular widths, starting at 52 pixels (5.7 octaves) between the eyes and increasing through 64 pixels up to 76 pixels (6.3 octaves) between the eyes. We then evaluate how well the Octave 6 MOSSE filter localizes faces for every scale of training image. The result is shown in Figure~\ref{octave6plot}.

The horizontal axis of Figure~\ref{octave6plot} is the difference in scale (measured in octaves) between the filter and test image. The vertical axis is the percentage of time the filter localizes the face to within 5 pixels of its true position (a tight threshold, considering the interocular distance is 64 pixels). The curve of the data indicates that when the scale mismatch between the filter and target is plus or minus a tenth of an octave, the ability of the filter to localize the target drops about 10\%. At plus or minus a sixth of an octave, localization ability drops about 20\%. Beyond that, the decline is rapid. This suggests that between 3 and 5 MOSSE face filters are needed to cover one octave in scale space.

The results in Figure~\ref{octave6plot} were calculated at octave 6. As shown in Figure~\ref{filters}, MOSSE filters are more detailed at larger scales and less detailed at smaller scales. To see if this makes a difference, we repeated the experiment above at interocular distances of 16, 32, 48, 64, 80 and 96 pixels (4, 5, 5.58, 6, 6.32 and 6.58 octaves, respectively). These results are presented in supplemental material and they are qualitatively similar. As a result, between 3 and 5 MOSSE filters are needed to cover an octave of test scales anywhere in scale space.

\subsection{Filter Selection}
\label{unprocessedTesting}
The previous experiment established that between 3 and 5 filters are needed per octave to compensate for changes in scale. Although the FERET faces are forward facing, when we move to more challenging domains we will also need multiple filters per scale to accommodate different poses. Using more than one filter introduces a challenge. Of the many filter responses to an image, how do we pick the local optima that matches best and therefore determines both the position and scale of the face? This section, establishes
a methodology to pick the best match filter responses. 


\subsubsection{Peak-to-Sidelobe-Ratio (PSR)}
To pick the best filter response from a batch of filter responses we use the Peak-to-Side-lobe Ratio (PSR) as recommended by Savvides et. al.~\cite{mace}. PSR is a probabilistic measure of the height of a peak in terms of the mean and standard deviation of the surrounding signal. The response peak with the highest PSR is selected as the best filter and used to determine the location and size of the face in the image.

The Peak-to-Side-lobe-ratio (PSR) is defined as:
\begin{equation}
PSR = \frac{peak - mean}{\sigma},
\end{equation}
where $peak$ is the value of a local maxima in the correlation surface, and, $mean$ and $\sigma$ are the mean and the standard deviation
of the correlation surface excluding a 5$\times$5 pixel window around the local maxima. 
It is pertinent to mention that Saviddes et al.~\cite{mace} did not compute the mean and standard deviation over the whole image (excluding the local peak). Instead, they estimated the mean and standard deviation in a 64$\times$64 pixel window around but excluding the local maxima. We tried this approach in pilot studies, but achieved better performance when the full image was used.

\subsubsection{Selecting Filter Responses}
The general approach to compensating for changes in scale and viewpoint in CorrFaD is to convolve the image with multiple filters and select the best responses. Unfortunately, more filters have the potential of introducing more false positives. To test this, we trained 13 filters at scales ranging from 3 octaves (8 pixels between the eyes) to 6.7 octaves (96 pixels between the eyes), and tested them on non-resized (unprocessed and random) test images. The results are presented in Table~\ref{newAcc}. One observation from this table is that most of the faces in the FERET data set have between 64 and 80 pixels between the eyes. Another is that the average face responds correctly to about three filters, reinforcing the observation that filters detect faces over a range of sizes.


The most important line in Table~\ref{newAcc} is the last one. This is the accuracy that results from selecting the peak with the highest PSR value for each image out of all the filter responses. This means that when an image is tested with each one of these filters, we pick the scale and pose of the filter that finds the highest PSR. Using this approach we correctly detected a face in 95.2\% of our test images. At this accuracy it is significantly higher than the accuracy of any of the individual filters, suggesting that the risk of additional false positives is more than compensated for by the ability to select among filters at different scales. 

\begin{table}[htp]
\caption{Accuracy of locating point between the eyes with filters trained on images with different number of pixels between the eyes and tested on unprocessed dataset.}
\begin{center}
\begin{tabular}{|c|c|}
\hline
 Training Pixel Octave &  \% Accuracy (within 5 pixels)\\
 \hline
3 &  0.0\\
\hline
4 &  0.4\\
\hline
4.58 & 8.8\\
\hline
5 & 2.8\\
\hline 
5.32 & 2.4\\
\hline
5.58 & 9.2\\
\hline
5.8 & 38.8\\
\hline
6 & 81.2\\
\hline
6.17 & 85.2\\
\hline
6.32 &  62.4\\
\hline
6.46 & 38.4\\
\hline
6.58 & 29.2\\
\hline
6.7 & 18.8\\
\hline
\hline
Max PSR & 95.2\\
\hline
\end{tabular}
\end{center}
\label{newAcc}
\end{table}%

\section{CorrFaD: Detecting Faces in Repeated Settings}
\label{corrfad}

As mentioned in the introduction, face detection in easy data sets like FERET is a solved problem. Most recent research concentrates on face detection in unconstrained settings, even if the resulting algorithms are computationally expensive. This paper develops a simple algorithm for detecting faces in complex but repeated settings, a situation common in many surveillance applications.

Specifically, we advocate detecting faces by convolving the source image with a set of linear filters, and thresholding the maximum response. This is such a simple (and old) technique that its use in complex face detection scenarios may be surprising. Two factors have changed, however: (1) new optimal filter techniques create linear filters that are more robust to changes in lighting, expression and other factors; and (2) MOSSE filters in particular learn to discount repeated backgrounds. As a result, faces can be detected in even complex scenarios as long as the general location is repeated.

\subsection{Datasets: PaSC and R-PaSC}
The Point-and-Shoot Challenge (PaSC) data set was introduced for the handheld video face and person recognition competition at International Joint Conference on Biometrics (IJCB) in 2014~\cite{ijcbpasc2014}. It was also the basis of a video person recognition evaluation at Face and Gesture conference in 2015~\cite{fgpasc2015} and video person recognition challenge at the $8^{th}$ IEEE International Conference on Biometrics: Theory, Applications, and Systems (BTAS) in 2016~\cite{pascChallenge2016}. The data set contains videos of people performing every day tasks. As a result, the people in the videos walk around through different lighting conditions, sometimes getting closer to the camera and other times getting farther away. They often change expressions, and they rarely look at the camera. In addition, the cameras are inexpensive point-and-shoot that are generally held by hand, resulting in jitter and motion blur. In other words, PaSC is a challenging data set near the difficult end of the face detection spectrum, similar in many ways to JANUS~\cite{klare2015pushing}. 

The PaSC data set is, in fact, too difficult for the linear filters advocated here. As shown below, they are not able to reliably detect faces in the PaSC data set. Unlike JANUS, however, the PaSC data set was collected at a small set of indoor and outdoor locations. As a result, it is possible to create a set of smaller, restricted data sets that we will collectively call R-PaSC (Restricted-PaSC). Every data set in R-PaSC contains a set of training videos that were all collected at one location. The associated test videos were collected at the same location. The people are disjoint across the training and test sets, so that we are never testing on a video of a person that we trained on. Other than the location, all other sources of variation (e.g. expression, task) remain.However, as we will show below, just restricting the training and test location simplifies this task enough that MOSSE filters are now able to detect the faces. 

To be specific, the unrestricted PaSC data set contains 2,802 videos of 265 people carrying out simple tasks. There are four to seven videos per subject using controlled videos and four to seven videos using hand held cameras in six different locations. 

\begin{figure*}[htbp!]
 \centering
 \subfigure[Image 1]{
 \label{im1}
 \includegraphics[width=2.3in]{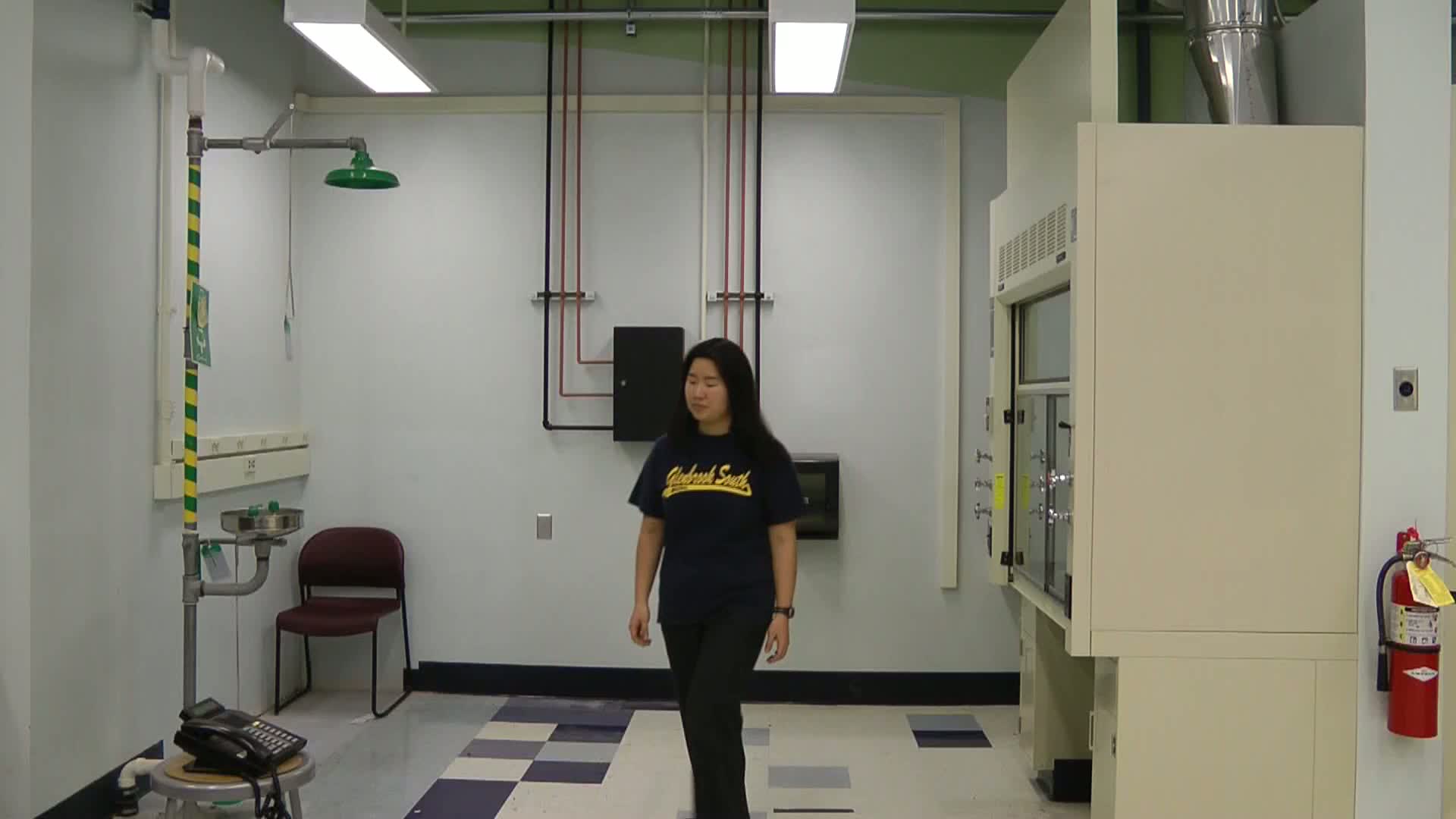}
 }
 \subfigure[Image 2]{
 \label{im2}
 \includegraphics[width=2.3in]{Images/06346d170-032.jpg}
 }
 \subfigure[Image 3]{
 \label{im3}
 \includegraphics[width=2.3in]{Images/06332d143-185.jpg}
 }
 \subfigure[Image 4]{
 \label{im4}
 \includegraphics[width=2.3in]{Images/05450d1360-010.jpg}
 }
 \caption{Representative images from the R-PaSC data set.}
 \label{rpasc}
\end{figure*}

Since our goal for R-PaSC is to restrict to a particular setting, out of the 2,802 unrestricted PaSC videos, we picked 256 videos shot at a particular location for training and 73 videos for testing. Figure~\ref{rpasc} displays frames from the videos shot at a particular location with different background objects. The details of this data set are presented in Table~\ref{rpasctable}. There is no overlap between the test and the training sets including people. The background is different in terms of the number of background objects but the location remains the same.

\begin{table}[htp]
\caption{Details of R-PaSC data set.}
\begin{center}
\begin{tabular}{|c|c|}
\hline
Number of Training Videos &  256\\
\hline
Number of Training Frames &  18966\\
\hline
Number of Test Videos &  73 \\
\hline
Number of Test Frames &  1865\\
\hline
Number of People in Training Videos & 128\\ 
\hline
Number of People in Test Videos & 36 \\
\hline
Number of locations & 1 \\
\hline
\end{tabular}
\end{center}
\label{rpasctable}
\end{table}%

\subsection{Training \& Selecting Filters}
\label{ONF}
The goal of this section is to evaluate whether a bank of MOSSE filters can be used to detect faces in surveillance applications with repeated settings. This requires that we first determine how many filters we need to span the range of scales and viewpoints we expect to encounter. Based on the available training data, we decide to limit the experiment to recognizing images of faces with between 16 and 128 pixels between the eyes (a three octave range) and in which both eyes are visible (i.e. no profiles or backs of heads). 

In Section~\ref{sec:feretscale}, we determined that between 3 and 5 filters per octave were needed to cover changes in scale. This was determined based on experiments with the FERET data set, so that confounding factors such as changes in illumination or expression would not alter the result. Real applications include these other factors, however, so we decided to confirm whether 3 to 5 filters per octave was still sufficient for this new, harder data set.

\begin{figure}[htbp!]
\begin{center}
\includegraphics[width=3.2in]{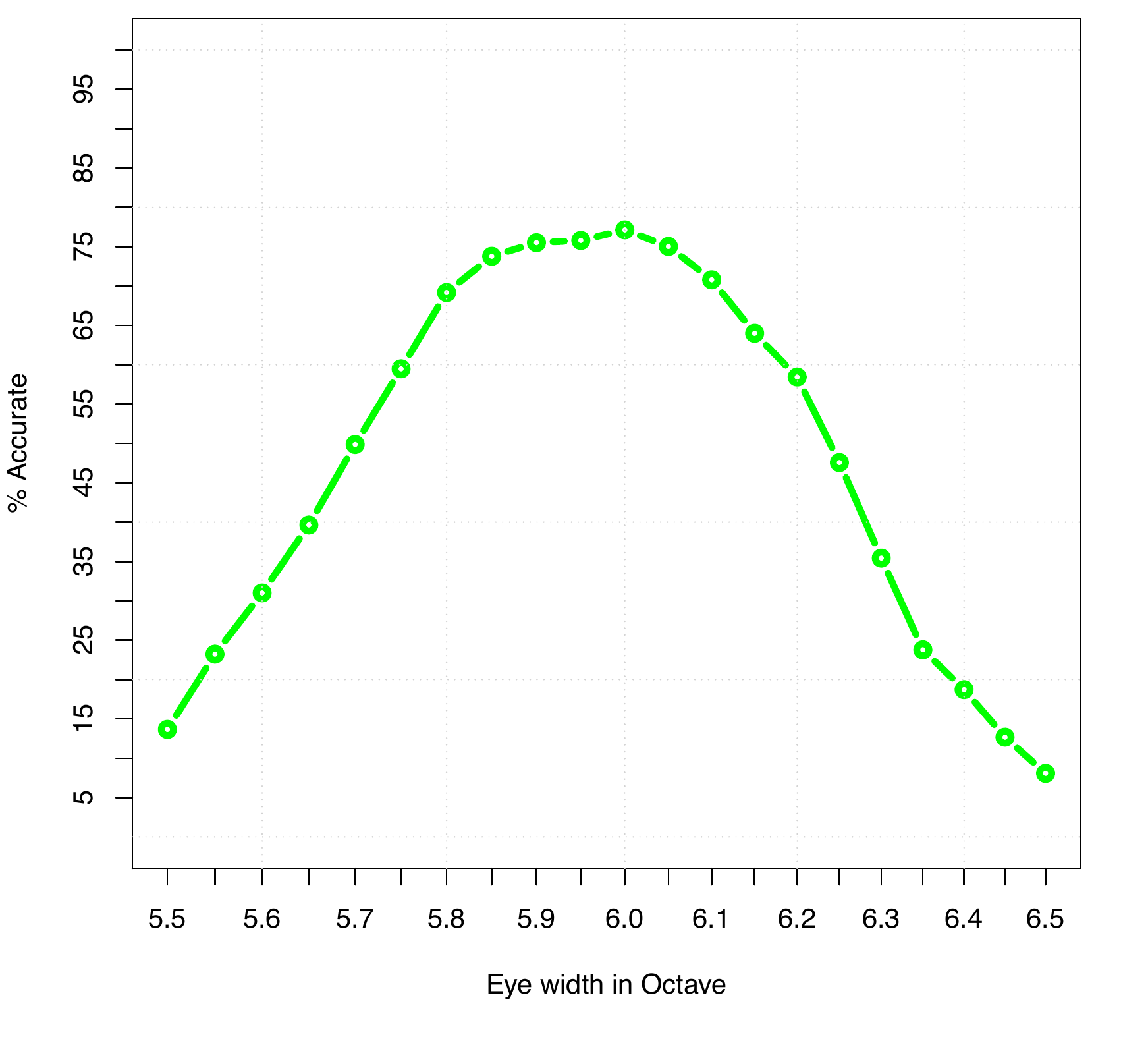}
\caption{{The percentage accurate is defined as the number of images which locate the face within
10\% of the eye width from the target (point between the eyes) location.}}
\label{optFilters}
\end{center}
\end{figure}

We ran a controlled experiment using the same methodology as in Section~\ref{sec:feretscale}. We resampled the training images to have 64 pixels between the eyes, and resampled the test images to have between 45 and 90 pixels between the eyes (i.e. plus or minus half an octave) in increments of 0.05 octaves. The resampling was done on test images from the PaSC data set that have an interocular width greater than or equal to 64 pixels to avoid "up sampling". The results are shown in Figure~\ref{optFilters}. 


Figure~\ref{optFilters} shows the accuracy of localizing a target within 10\% of the inter-ocular width in both x and y directions. Specifically, it shows the results of
a filter trained for images of octave 6 inter-ocular width (i.e. 64 pixels) when tested on 21 different data sets whose inter-ocular widths range between 5.5 octaves and 6.5 octaves. 

According to Figure~\ref{optFilters}, the detection accuracies for the quarter octave between
5.86 and 6.10 octaves stay between 70\% and 75\%. This is consistent with the previous results on FERET, and lead us to train four MOSSE filters per octave.
We confirmed this result with additional experiments for filters trained on other 
inter-ocular widths. The results were consistent across these additional tests
The next challenge is to determine how many filters we need per scale to account for changes in viewpoint. We ran a pilot experiment to determine the range of pose variation, and found that there are three different ranges
within the PaSC training dataset: less than -12${^\circ}$ (left side facing the camera), between -12${^\circ}$ and +12${^\circ}$ (in and around frontal pose) and greater than +12${^\circ}$(right side facing the camera). So, we have frontal or near frontal pose, left
profile or between frontal and left profile pose and right profile or between right profile and near frontal pose. As a result, we use train filters for four scales and three viewpoints per octave, for a total of 12 filters per octave. Since we want to cover a range of interocular distances ranging from 16 to slightly over 128 pixels, we train a total of 39 filters covering $3\frac{1}{4}$ octaves.

\begin{figure*}[htbp!]
\centering
\subfigure[pose \textgreater 12$^\circ$]{
\label{scale32im1}
\includegraphics[width=2.2in]{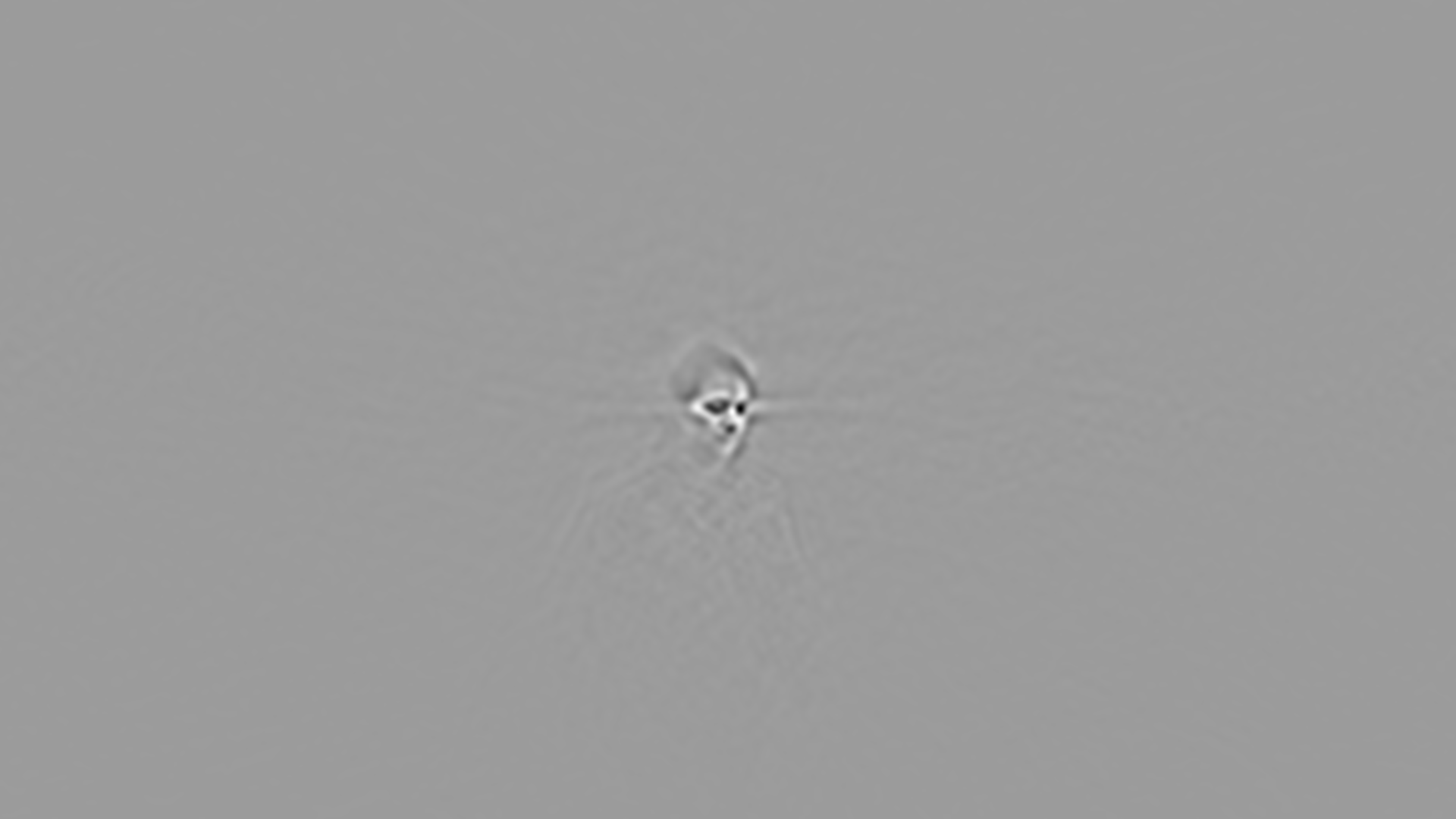}
}
\subfigure[pose between -12$^\circ$ and +12$^\circ$]{
\label{scale32im2}
\includegraphics[width=2.2in]{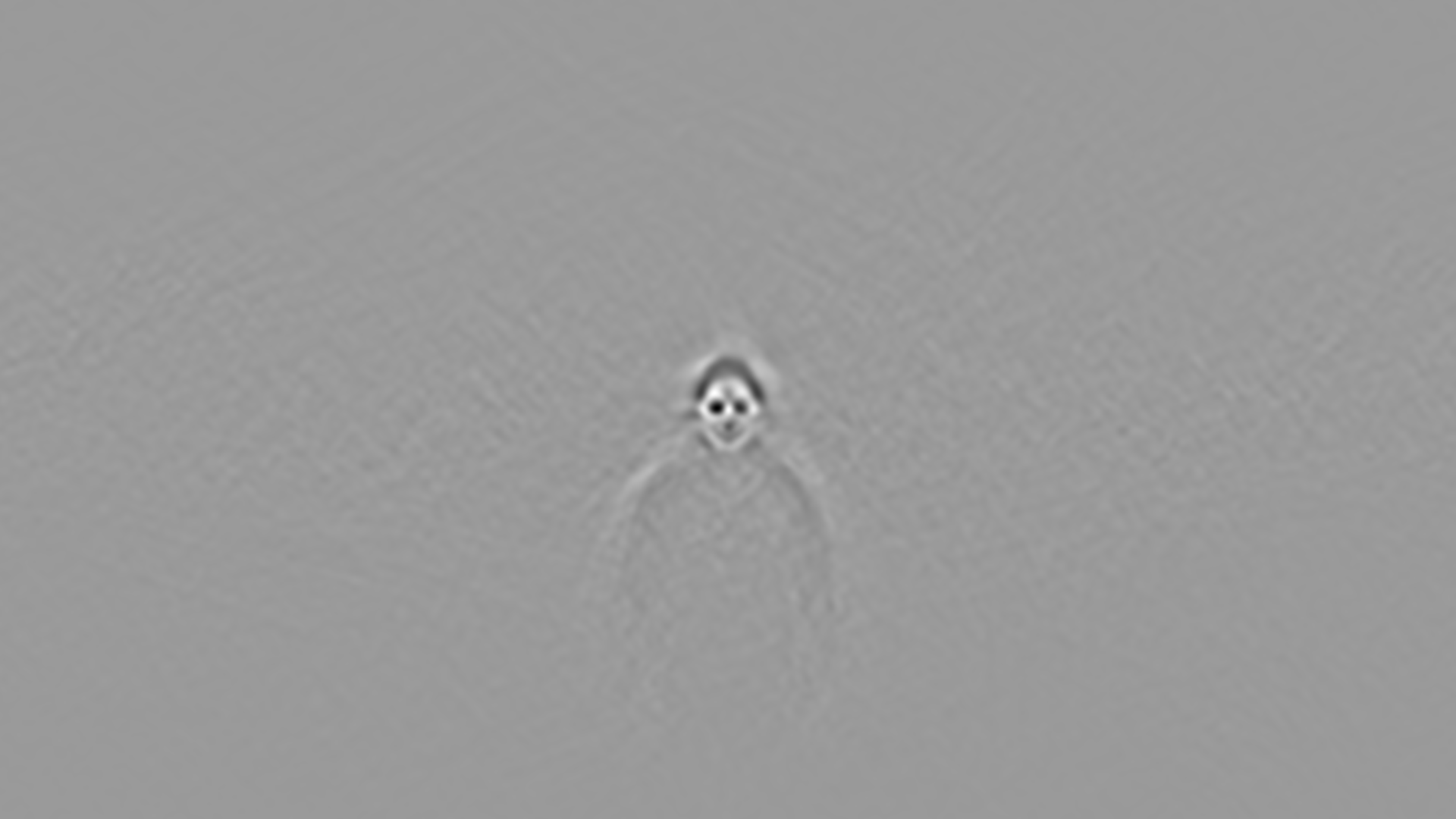}
}
\subfigure[pose \textless 12$^\circ$]{
\label{scale32im3}
\includegraphics[width=2.2in]{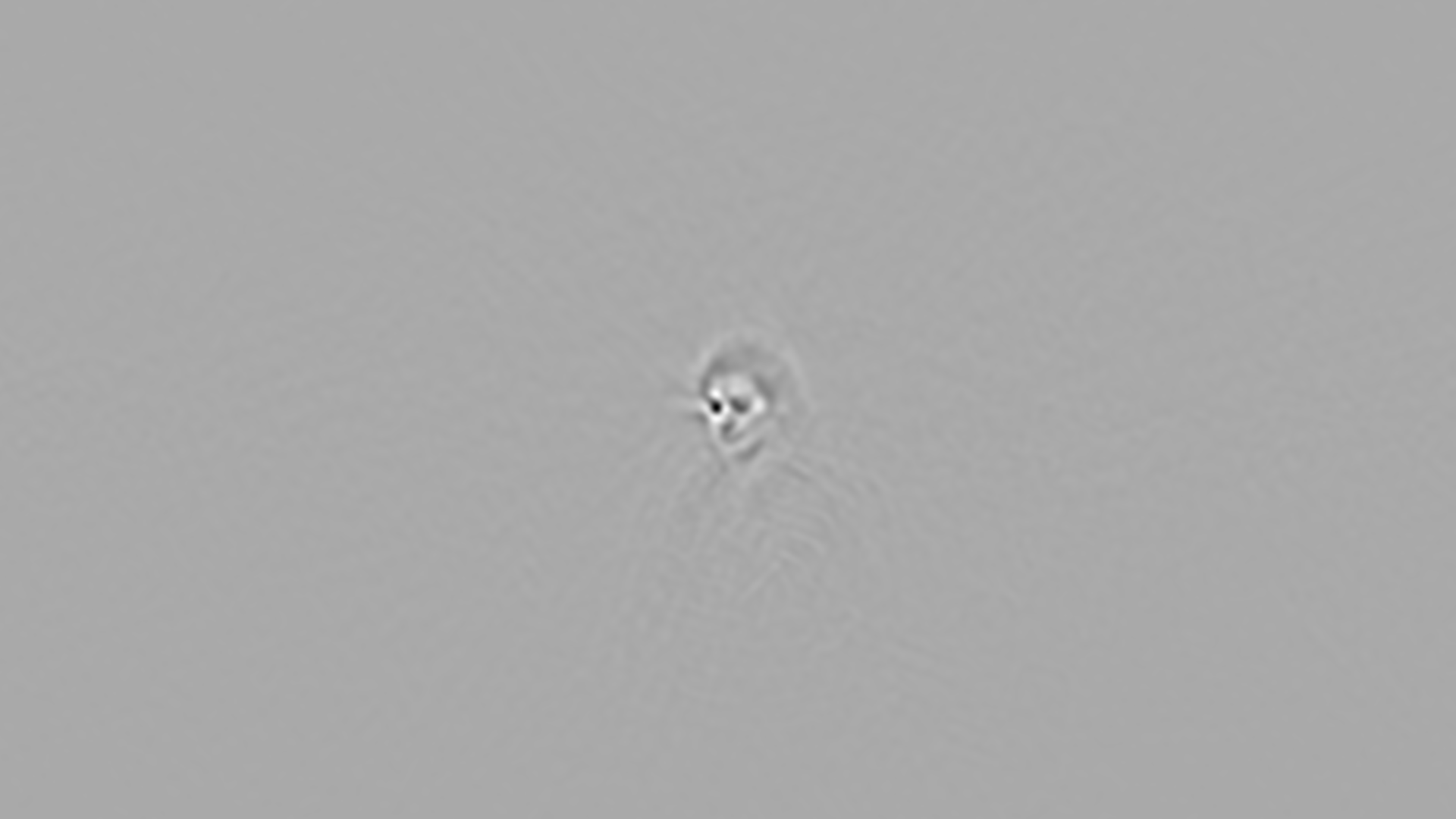}
}
\caption{Three different poses for a scale of 32 pixels between the eyes.}
\label{scale32}
\end{figure*}
\begin{figure*}[htbp!]
\centering
\subfigure[pose \textgreater 12$^\circ$]{
\label{scale64im1}
\includegraphics[width=2.2in]{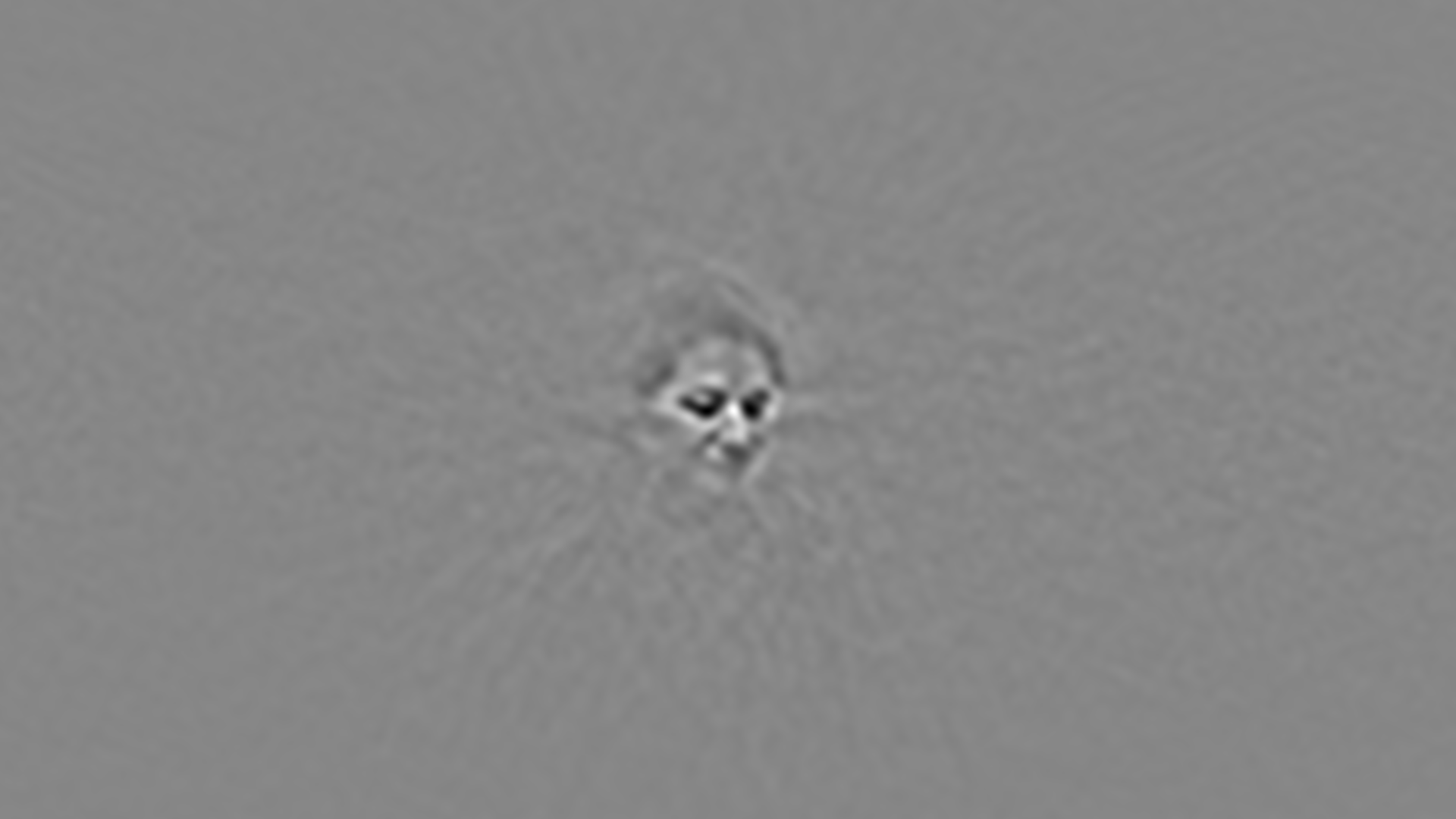}
}
\subfigure[pose between -12$^\circ$ and +12$^\circ$]{
\label{scale64im2}
\includegraphics[width=2.2in]{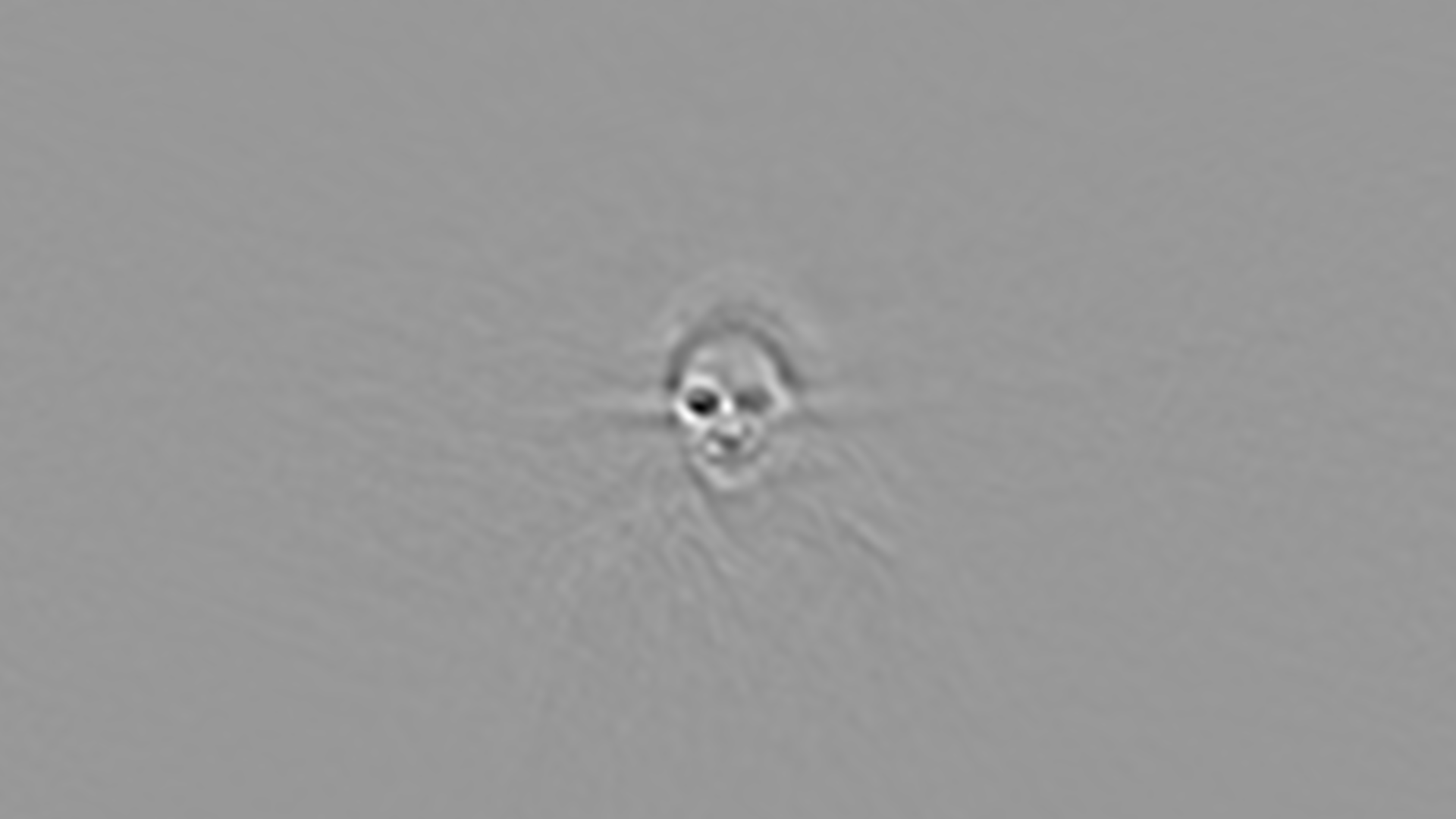}
}
\subfigure[pose \textless 12$^\circ$]{
\label{scale64im3}
\includegraphics[width=2.2in]{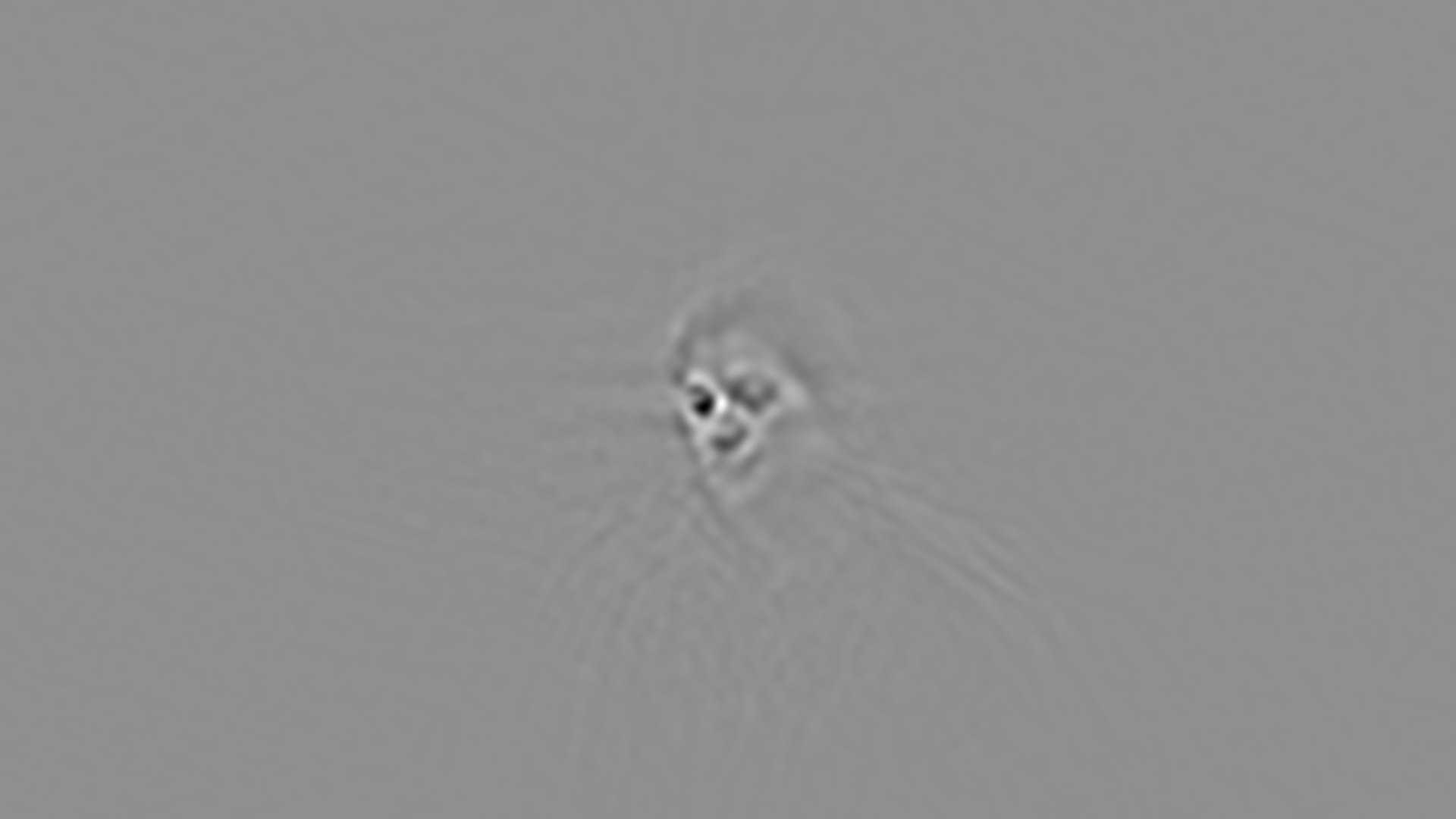}
}
\caption{Three different poses for a scale of 64 pixels between the eyes.}
\label{scale64}
\end{figure*}
\begin{figure*}[htbp!]
\centering
\subfigure[pose \textgreater 12$^\circ$]{
\label{scale90im1}
\includegraphics[width=2.2in]{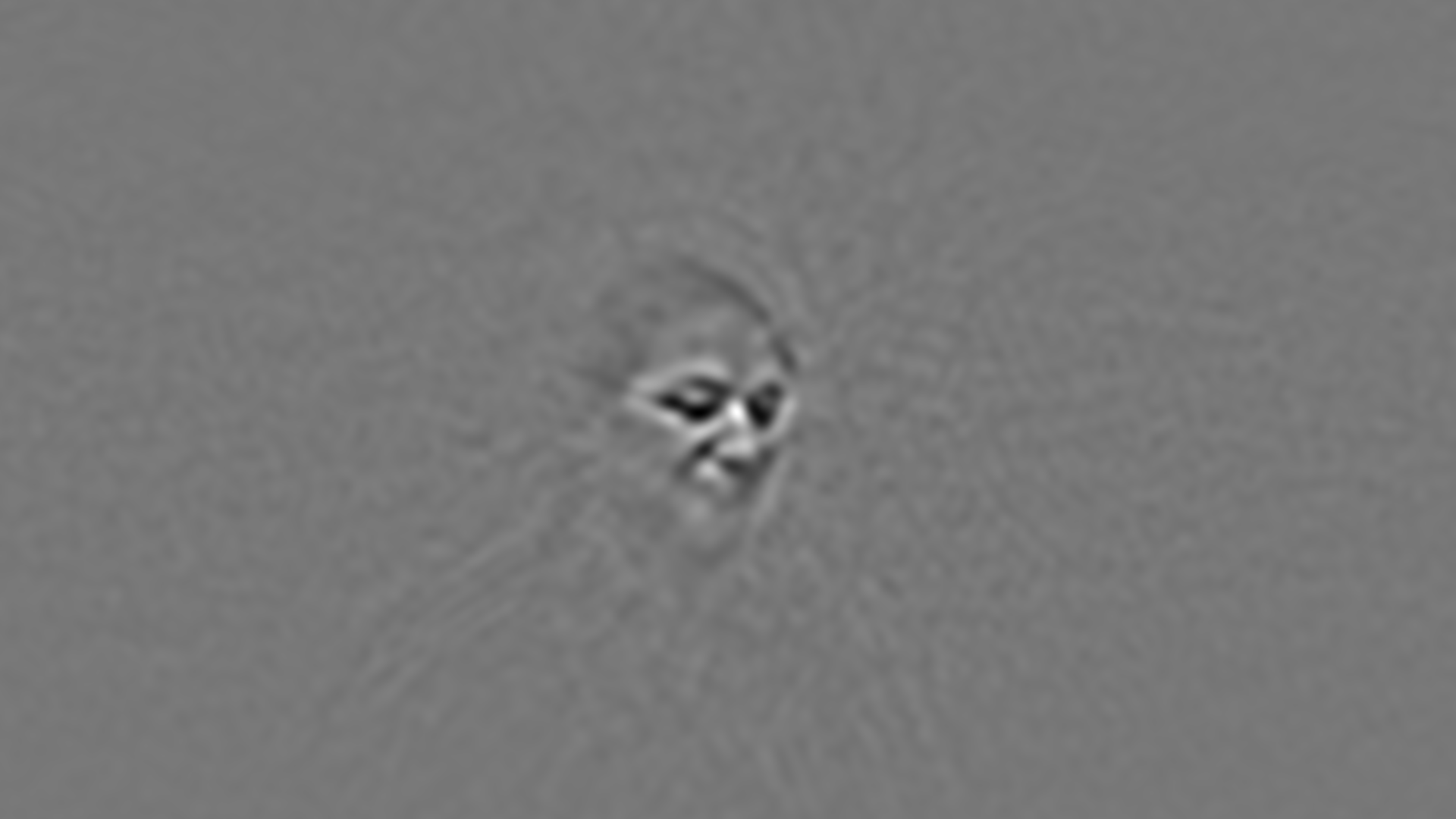}
}
\subfigure[pose between -12$^\circ$ and +12$^\circ$]{
\label{scale90im2}
\includegraphics[width=2.2in]{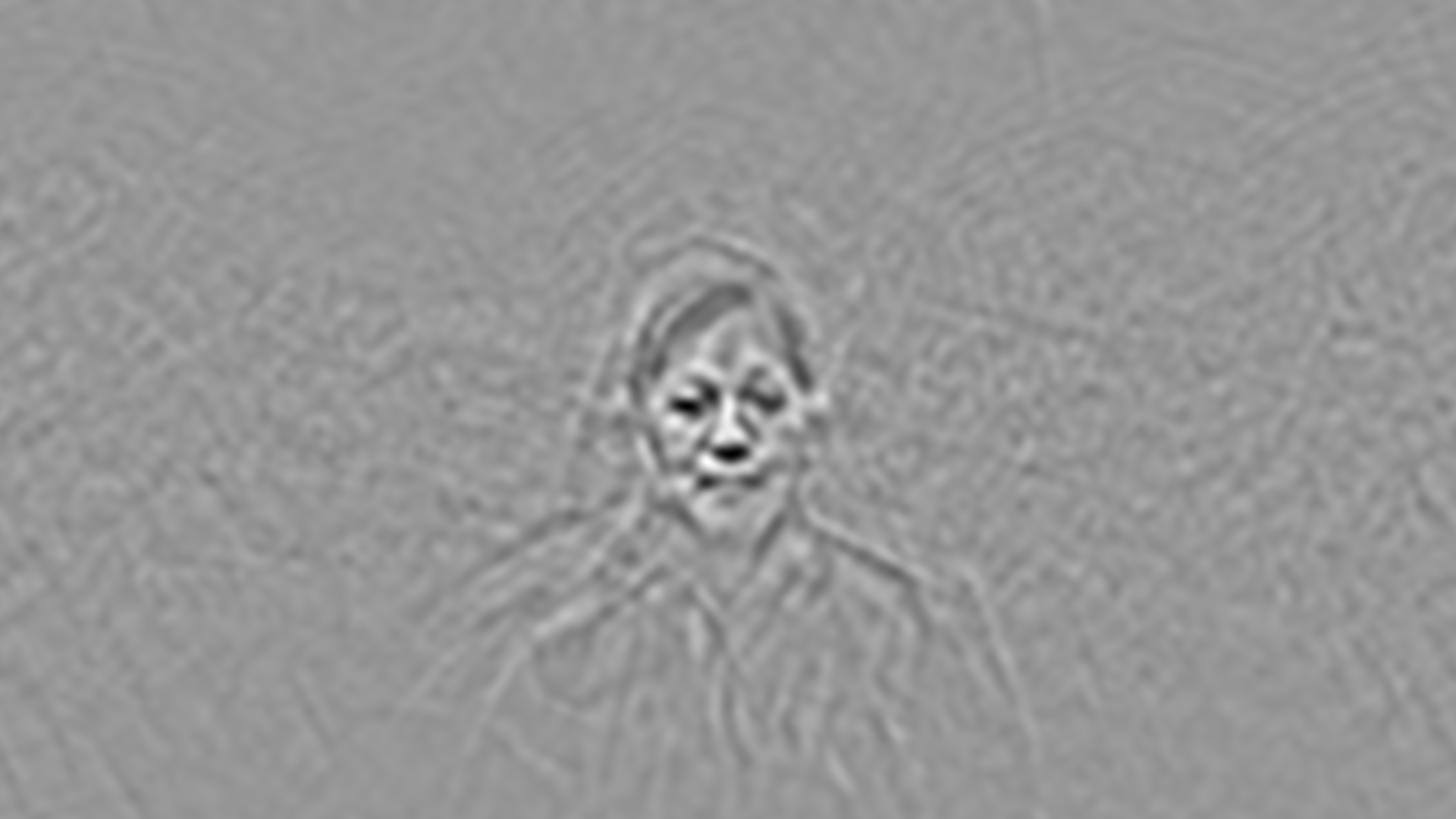}
}
\subfigure[pose \textless 12$^\circ$]{
\label{scale90im3}
\includegraphics[width=2.2in]{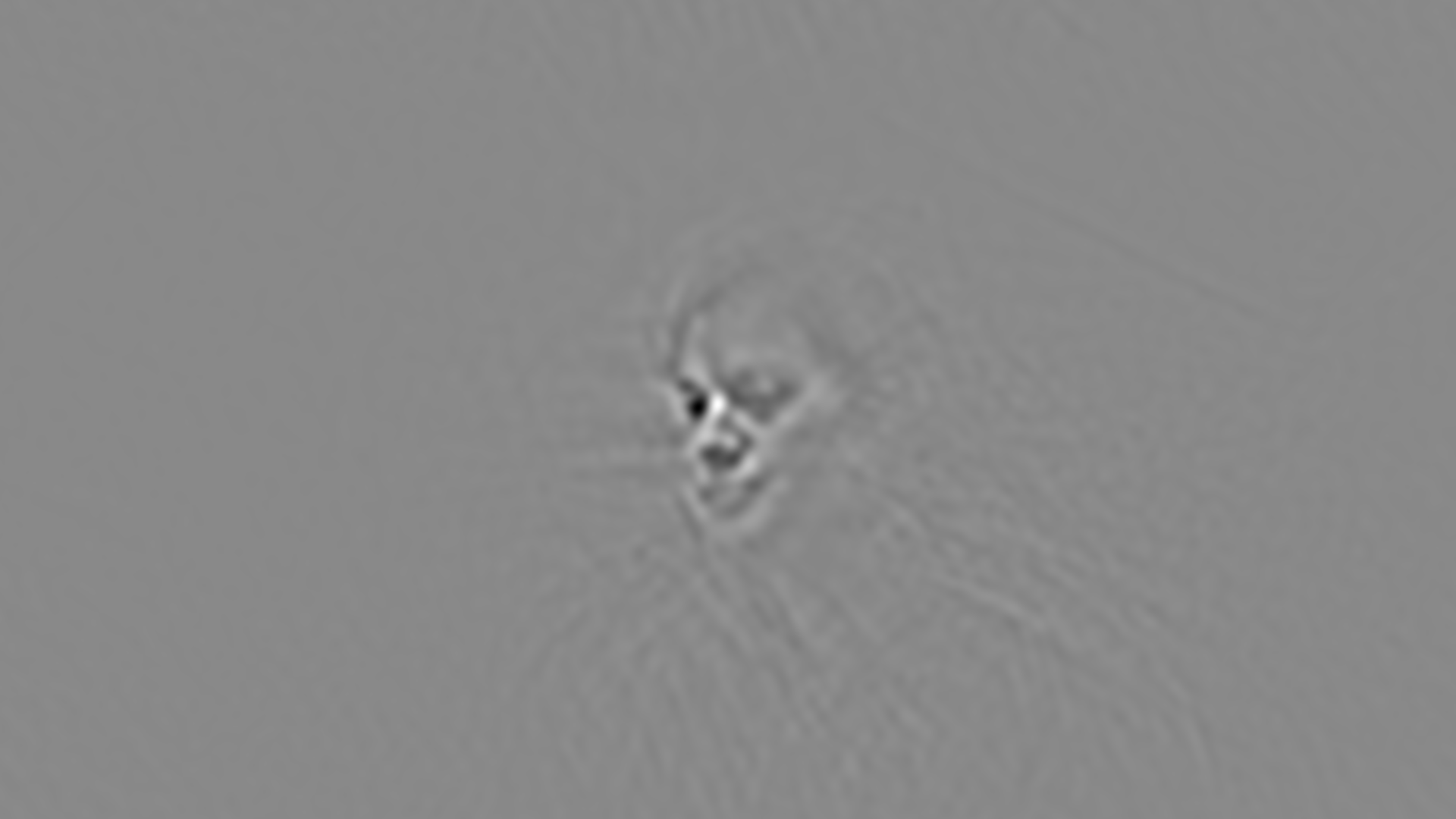}
}
\caption{Three different poses for a scale of 90 pixels between the eyes.}
\label{scale90}
\end{figure*}

When training these filters, we did not artificially resize any face images. Instead, we used 255,835 frames selected from 999 training videos to cover all 39 combinations of pose and scale. Nine of those trained filters are displayed in
Figures~\ref{scale32},~\ref{scale64} and~\ref{scale90}. The three filters
in each figure have the same scale but different poses. To be more precise, in Figure~\ref{scale32}, the left filter~\ref{scale32im1} has been trained on an image dataset containing images with 32 pixels between the eyes and a pose of
greater than 12$^\circ$ or the right profile. Similarly, the filter in Figure~\ref{scale32im2} is trained on a dataset with
images that have 32 pixels between the eyes and a frontal or near frontal pose. Finally, Figure~\ref{scale32im3}
shows a filter trained on a dataset having images that have 32 pixels between the eyes and a pose of
less than 12$^\circ$ or near left profile. Figures~\ref{scale64} and~\ref{scale90} can be explained in a
similar fashion. 

\subsection{Evaluation Methodology}
To evaluate the bank of MOSSE filters selected above as a face detection mechanism, we use two approaches. 
The first approach to find faces in this
test dataset was the same as the bank of filters approach of Section~\ref{unprocessedTesting}. However,
after many experiments it started to become clear that the convolution of the filters with a test image in the
Fourier domain showed very poor accuracy results. On the contrary, our second approach, using spatial convolution to match just the faces with the test image proved to be a better option as is discussed in the next section.
\subsubsection{Convolution Using Spatial Correlation}
In this section, we present a case for using a bank of filters approach in the spatial domain instead of the bank of
filters approach in the Fourier domain because the former approach is more effective on the PaSC dataset
than the latter one. Although it seems contrary to some of my earlier approaches, a set of careful experiments
clearly made it an approach of choice for this dataset. One of the
preprocessing steps before the convolution in the Fourier domain involves normalization of the full image. This increases the likelihood of a higher correlation value at a location other than on the face, for example, near a
light source. On the contrary, normalization in the case of spatial correlation is applied for each specific location
separately, corresponding to the size of the template, which is usually much smaller than the image size. This
approach gives a better chance of a correct match.

\begin{figure}[htbp!]
\begin{center}
\includegraphics[width=3in]{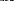}
\caption{{Input Image}}
\label{testInpImage}
\end{center}
\end{figure}

\begin{figure}[htbp!]
\begin{center}
\includegraphics[width=1in]{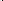}
\caption{{Template Image}}
\label{testtemplate}
\end{center}
\end{figure}

\begin{figure}[htbp!]
\begin{center}
\includegraphics[width=3in]{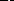}
\caption{{Template Image of Figure~\ref{testtemplate} with Padded zeros}}
\label{templateWithZeros}
\end{center}
\end{figure}

\begin{figure}[htbp!]
\begin{center}
\includegraphics[width=3in]{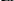}
\caption{{Correlation Output after convolving template Image of Figure~\ref{templateWithZeros} with Padded zeros and Figure~\ref{testInpImage}. }}
\label{corrEx}
\end{center}
\end{figure}

In order to explain the difference between the two approaches a little more, we will now use a toy example.
Consider Figure~\ref{testInpImage} to be an input image and Figure~\ref{testtemplate} to be the filter (or template).
We are interested in finding this template in the input image.
The first step is to normalize the input test image before convolving it with the template in the Fourier domain.
Secondly, since the template is not the same size as the test image, the template needs to be padded with zeros to
make it the same size as the test image. The template with the padded zeros is shown in Figure~\ref{templateWithZeros}.
Convolution of the template and the test image results in a correlation surface with a maximum at coordinates $(0,0)$. This correlation surface is displayed in Figure~\ref{corrEx}.
On the contrary, spatial correlation, using OpenCV template matching, correctly identifies the peak at coordinates $(8,0)$. For the rest of this chapter we will follow the spatial template matching approach.

Specifically, the testing of PaSC datasets was carried out by using OpenCV template matching~\cite{opencv_library}.
It is basically spatial correlation of templates with the test images.
In particular, we used the normalized cross correlation coefficient  of OpenCV template matching. It is mathematically
represented by the equation~\ref{ncceq}. 
\begin{strip}
\begin{equation}
NCC(x,y) =\frac{ \sum_{i=1}^{W}\sum_{j=1}^{H}I(x+i,y+j)\cdot T(i,j)}{\sqrt{\sum_{i=1}^{W}\sum_{j=1}^{H}I(x+i,y+j)^2}\cdot\sqrt{\sum_{i=1}^{W}\sum_{j=1}^{H}T(i,j)^2}},
\label{ncceq}
\end{equation}
\end{strip}
where $I$ is a source image, $T$ is a matching template of size $W \times H$, and $(x,y)$ is the location on the
image, $I$, where the template is centered for matching. 

For our experiments on the PaSC dataset, the templates to match were created by cropping out the faces from the trained filters. Since there are 39 filters corresponding to 13 scales and 3 poses for each scale, there are as many face templates. Some of these templates, displayed below in Figures~\ref{filter16} through ~\ref{filter32} for scales of 16 through 32 pixels(4-5 octave in quarter steps) between eyes, display a pyramid of filters of different scales between these octaves.
\begin{figure}[htbp!]
\centering
\subfigure[\textgreater 12$^\circ$]{
\label{im1}
\includegraphics[]{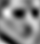}
}
\subfigure[\textgreater  -12$^\circ$ and \textless +12$^\circ$]{
\label{im2}
\includegraphics[]{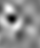}
}
\subfigure[\textless -12]{
\label{im3}
\includegraphics[]{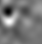}
}\\
\caption{filter templates for 16 pixels (octave 4) between the eyes.}
\label{filter16}
\subfigure[\textgreater 12$^\circ$]{
\label{cor1}
\includegraphics[]{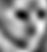}
}
\subfigure[\textgreater  -12$^\circ$ and \textless +12$^\circ$]{
\label{cor2}
\includegraphics[]{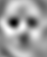}
}
\subfigure[\textless -12]{
\label{cor3}
\includegraphics[]{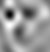}
}\\
\caption{filter templates for 19 pixels (octave 4.25) between the eyes.}
\label{filter19}
\subfigure[\textgreater 12$^\circ$]{
\label{im1}
\includegraphics[]{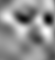}
}
\subfigure[\textgreater  -12$^\circ$ and \textless +12$^\circ$]{
\label{im2}
\includegraphics[]{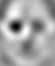}
}
\subfigure[\textless -12]{
\label{im3}
\includegraphics[]{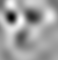}
}\\
\caption{filter templates for 22 pixels (octave 4.5) between the eyes.}
\label{filter22}
\subfigure[\textgreater 12$^\circ$]{
\label{cor1}
\includegraphics[width=1in,height=1in]{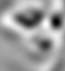}
}
\subfigure[\textgreater  -12$^\circ$ and \textless +12$^\circ$]{
\label{cor2}
\includegraphics[]{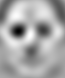}
}
\subfigure[\textless -12]{
\label{cor3}
\includegraphics[]{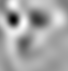}
}\\
\caption{filter templates for 26 pixels (octave 4.75) between the eyes.}
\label{filter26}
\subfigure[\textgreater 12$^\circ$]{
\label{im1}
\includegraphics[scale=0.9]{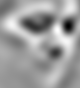}
}
\subfigure[\textgreater  -12$^\circ$ and \textless +12$^\circ$]{
\label{im2}
\includegraphics[scale=0.9]{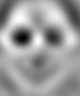}
}
\subfigure[\textless -12]{
\label{im3}
\includegraphics[scale=0.9]{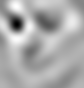}
}\\
\caption{Filter templates for 32 pixels (octave 5) between the eyes.}
\label{filter32}
\end{figure}

The approach to detect faces in still images and the frames in videos is carried out using each of these filter templates, called the bank of filters, to match with a test image. Each of these templates is matched with a given
test image using OpenCV template matching. The location of the peak correlation value is recorded. This
location combined with the filter width and height forms a rectangle representing the detected face in the test image.  A match is considered to be found between the returned face rectangle and the ground truth face rectangle if there is an overlap of 25\% between the two rectangles. The ground truth face rectangles were
obtained using the SDK 5.2.2 version of an algorithm developed by Pittsburgh Pattern Recognition (PittPatt)~\cite{pasc}.

\section{Location Specific Face Detection}
\label{ssfd}
The focus of this research is to detect faces in places, such as entrance to a sensitive building like an airport, nuclear facility, or any other places where confidentiality matters. With the goal of verification, the location in these settings remains more or less uniform with no control on pose or scale. In this section we present our results on a customized data set, obtained from the PaSC data set to maintain a uniform background. We hypothesize, a correlation filter designed for a specific setting with a specific background and not controlled for scale and pose, will yield a high accuracy in face detection. Rest of this section is dedicated to test this hypothesis.
\subsection{Experimental Setup}
This experiment is setup to test the hypothesis presented above. We have controlled
for location and as a result created a customized dataset of video frames. In this
dataset the location is the same between the training and the test datasets but there is no overlap of people
between the two datasets. All the videos have been shot in an indoor location. Some of the frames from six
different videos are presented in Figure~\ref{customFrames}.

\begin{figure*}[htbp!]
\subfigure[Filter with scale of 16 pixels between the eyes.]{
\label{fil16}
\includegraphics[width=2.45in]{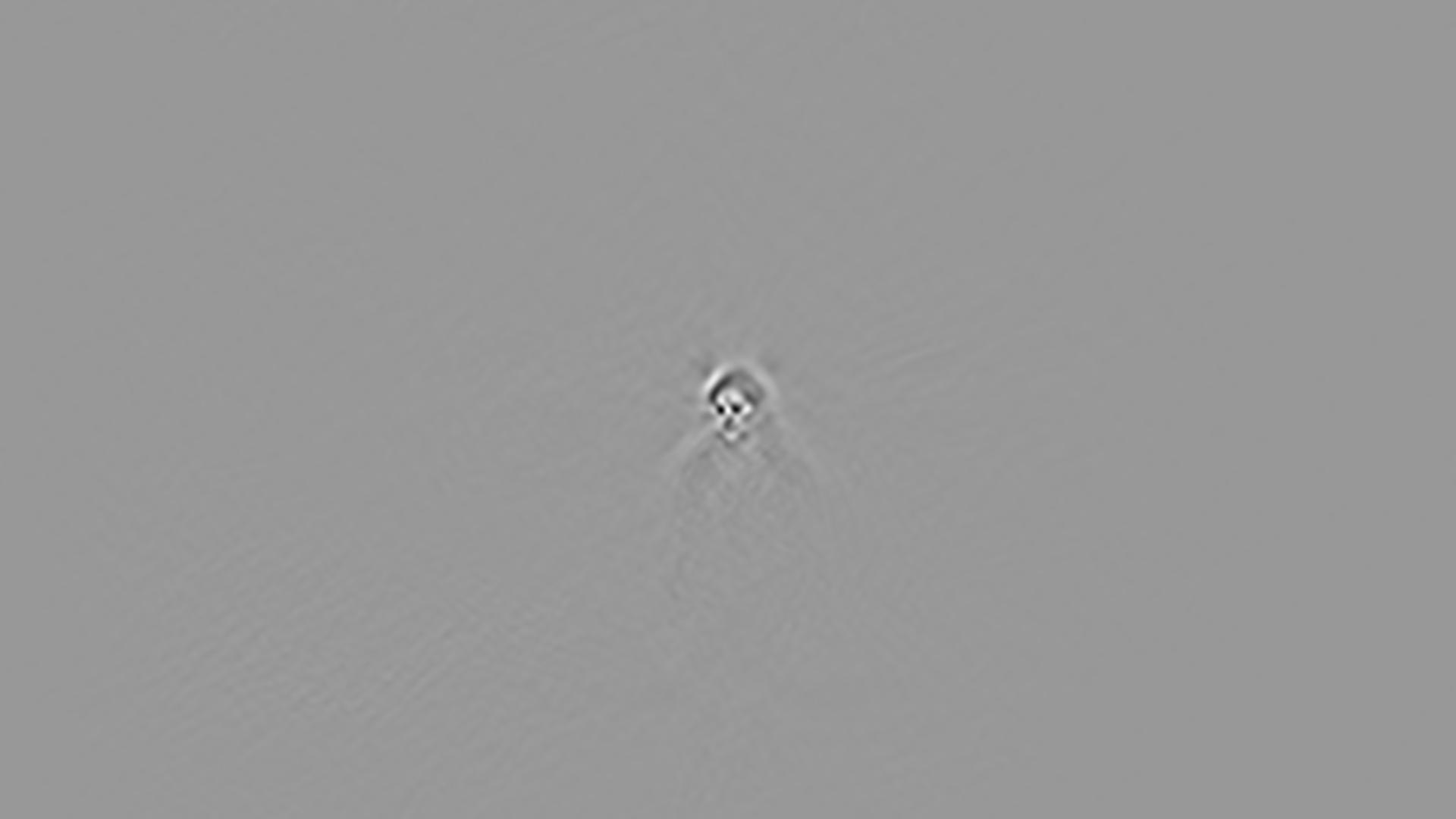}
}
\subfigure[Template (filter) cropped from ~\ref{fil16}.]{
\label{face16}
\includegraphics[]{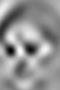}
}
\subfigure[Filter with a scale of 24 pixels between the eyes.]{
\label{fil24}
\includegraphics[width=2.45in]{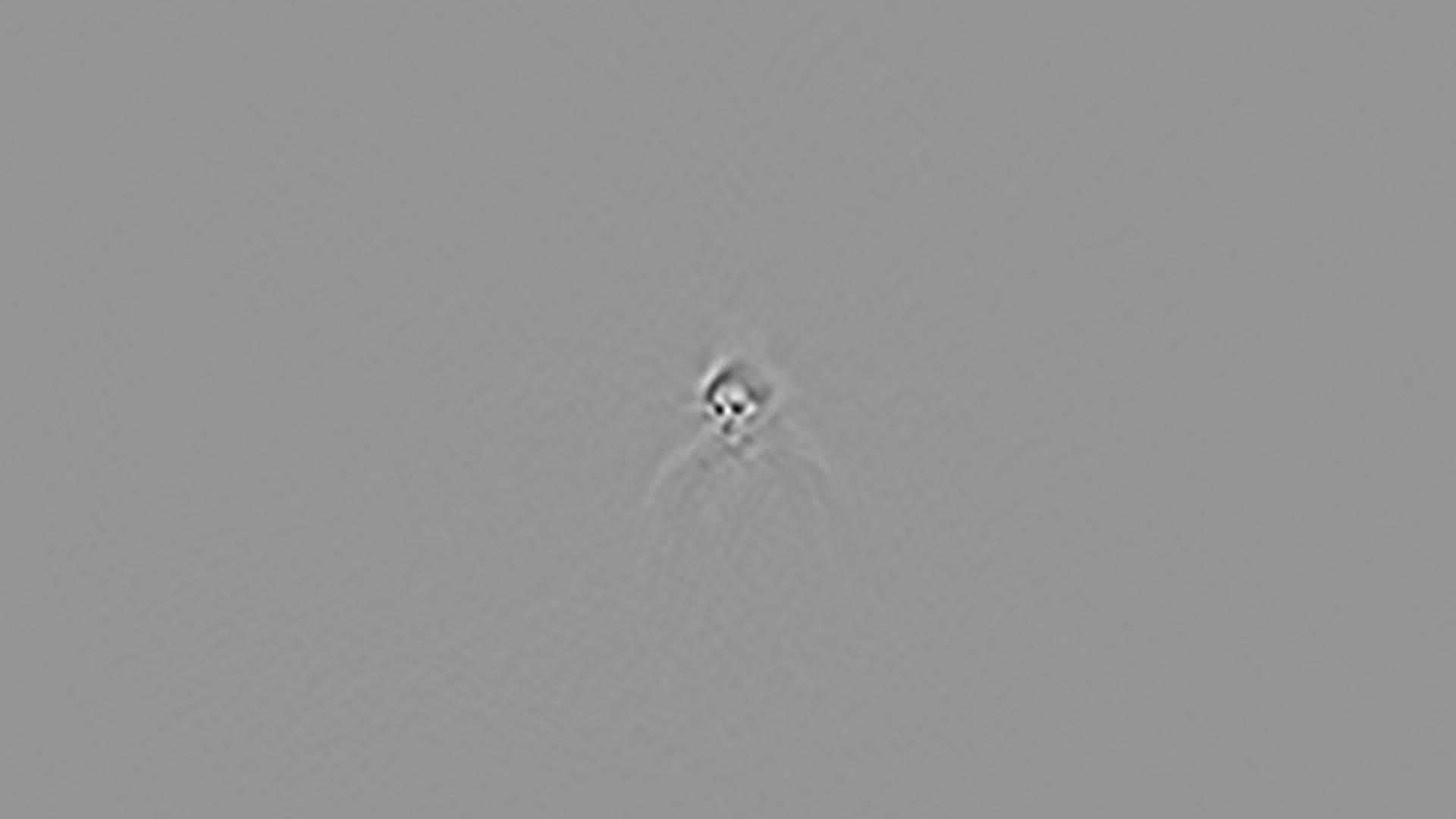}
}
\subfigure[Template (filter) cropped from ~\ref{fil24}.]{
\label{face24}
\includegraphics[]{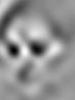}
}
\caption{Filters trained on 256 videos with 18966 frames for a particular location and two different scales.}
\label{customFilters}
\end{figure*}

These frames are a representative of this dataset displaying some of the variations in scale and slight pose,
in a specific setting. The training set consists of 256 videos with 18,966 frames. The filters
trained for this particular location, and two closely related poses, and two different scales
are shown in Figure~\ref{customFilters}. 

The faces in the training video frames for this setting have an inter-ocular width between 16 and 32 pixels and two closely related
poses near the left profile. Therefore, we created two different datasets for training. One set consists of images
having greater than or equal to 16 but less than 24 pixels between the eyes. This training
set is used to train the filter shown in Figure~\ref{fil16}. The second training set consists of images that had an
interocular width between 24 and 32 pixels. The filter trained on this dataset is presented in Figure~\ref{fil24}.

Even though the filters are trained on the full frames to get rid of the clutter in the background, smaller face
templates, shown in Figures~\ref{face16} and ~\ref{face24}, have been cropped from these trained filters to do a
spatial template matching using OpenCV. This ensures that the normalization is taken at each location of an
image corresponding to the size of the face template and not the entire image. 

These two face templates have been spatially correlated, using OpenCV, with the frames of a test set consisting
of 73 videos, containing 1,865 frames. Each of these videos has been shot at the same
location and the inter-ocular width varies between 18 and 33 pixels. It may be reiterated that
in our setup two filters have been trained on images with an interocular range of 16 to 32 pixels. If we take a look at
Figure~\ref{optFilters}, one may notice that there is a range of test images that would give a good detection accuracy
for a filter trained on images that are closer to the interocular width in the test dataset. Since the interocular
widths of the images in the training and test datasets lie in the same range, the filter design choices in terms
of interocular distance are appropriate.

\subsection{Results}
Each frame in the test dataset is matched with the face templates of Figures~\ref{face16} and~\ref{face24}, and
the coordinates of the location that returns the highest correlation value for these two matches is recorded.
To choose between these two results, the coordinates of the one that has a higher correlation value is selected.
This coordinate identifies the top left corner of the detected face rectangle. The width and the height
of the face template determines the width and the height of the face rectangle. The face rectangle
coordinates for each test image are recorded. These face rectangles are eventually used to determine the
performance of the experiment.

The performance measure used in this experiment is based on an overlap of the face rectangle obtained
by template matching of the face filters and the test images, with the face rectangle of the
faces from the ground truth for each of these images. If there is an overlap of 25\% or more between the two
rectangles, the face in an image is recorded as detected. 

\begin{figure}[htbp!]
\begin{center}
\includegraphics[width=3.0in]{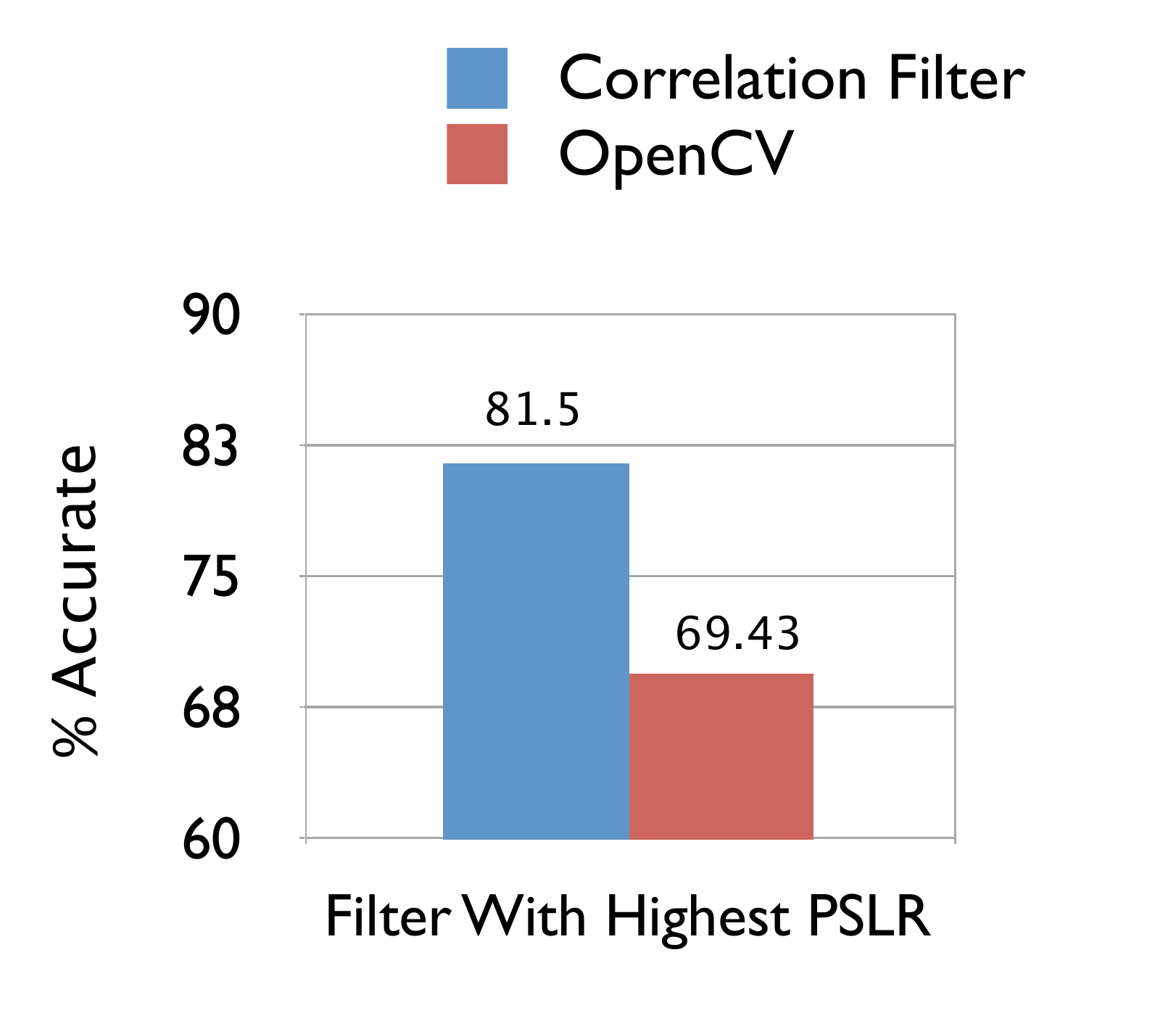}
\caption{{Comparison of Face detection accuracy between Correlation Filter approach and OpenCV Viola and Jones face detector.}}
\label{customVidRes}
\end{center}
\end{figure}

Based on this criteria for accuracy, the algorithm was
able to find a face match in 1,520  frames out of 1,865 frames which equates to an 81.5\% detection rate. This result has been compared with face detection results obtained from OpenCV Viola and Jones' face
detector, which achieved an accuracy of 69.43\% or 1,295 out of 1,865 frames. The criteria of accuracy for
this face detector is the same as for the correlation filter based face detector. These two results are displayed
side-by-side in Figure~\ref{customVidRes}. It is clear that the correlation based face detector outperforms OpenCV
Viola and Jones face detector in a specific scenario and it validates the hypothesis presented in the beginning of
this section.
\subsection{Conclusion}
In this section, we hypothesized that the more specific a filter, the better the accuracy in detecting faces in an image.
As such, a controlled experiment was devised in which the location in both training and test datasets was
restricted.
The inter-ocular widths between the training and the test datasets are also in the same range. The correlation
filter based face detector was used to test the dataset and compared with the OpenCV Viola and Jones face
detector. The results presented in Figure~\ref{customVidRes} clearly indicate that the correlation based approach
is doing better for such a dataset and hence confirms the hypothesis.

This is one of the most significant face detection results obtained using correlation filters. A constant
background setting gives an advantage to the approach because the correlation filter training appears to learn this
setting and returns a filter suited for a similar background. Because of this learning, the filters
ignore the background clutter during the testing process and avoid false positives. Therefore, one of
the most important application of a correlation based face detection approach is to use it in specific situations.
For example, people walking into a building through a specific door, or a security checkpoint where the setting remains mostly constant.
These results are important to establish the significance of face detection using correlation filters that
are trained for specific scenarios and is an attempt to lead this research in that direction.

\section{Limitations of CorrFaD}
Having established the scenario for which correlation based face detector, CorrFaD, is designed, this research would not be complete without studying the limitations of this approach. As previously admitted our approach is not designed for face detection in uncontrolled settings but rather repeated settings with a goal of face verification. In this section we summarize the findings of using CorrFaD in uncontrolled settings for images and videos (refer to supplemental material for a detailed discussion).

These image and video data sets represent a careful combination of uncontrolled scale, pose, lighting, and location along with the
sensors used to capture it, just like numerous, handheld cameras and cellphones that people around the world
use to take pictures.
The results have been compared with the random results as well as OpenCV Viola and Jones face detector. It
turns out that although our approach does much better than random experiment, it is not at all competitive with the
Viola and Jones' face detector when a single filter is used. However, when all thirty-nine filters covering thirteen
different scales and three poses for each scale, are used, the accuracy of face detection by at least one of those
filters comes very close to the Viola and Jones face detection results. To be precise it is 70.54\% accuracy for our
approach versus 87.31\% accuracy using Viola and Jones face detector, on the still image data set. On the video
data set, the accuracy is 84.07\% versus 86.24\%, for correlation filter based face detector and Viola and Jones
face detector, respectively.
The reason for presenting and comparing the results when all filters are being used is that even though, the filter
that gives the top correlation value does not always find the right face, there is at least one filter in the set, that detects the
face correctly with a high accuracy. This verifies the apprehension that correlation based face detection is not
an approach of choice for so many variations (scale, pose, background etc.,) in the dataset. In fact we do not present it as such either and position it as an approach to be used for controlled background settings.

\section{Discussion}
In this paper we introduce a novel correlation based face detector, CorrFaD. 
Using CorrFaD for face detection involves a transition from target localization to face detection, using a batch of filters to determine the scale and pose of faces and finally, applying peak-to-sidelobe ratio to determine the face rectangle. It was applied to FERET, and PaSC image and video data sets.

Each of these images and video frames contains only a single face. CorrFaD is easy to train and fast in testing. However, like any other face detector this one also has to deal with many complexities associated with face detection.
Some of these complexities are associated with faces of different sizes or scales, pose, uncontrolled lighting,
background and location, while some others arise out of the basic principles of the process on which such a face
detector is based.

Any correlation operation returns a peak value and hence by design it is suited to locate one face in an image associated with the highest peak. Our goal from the outset has been to work and apply this algorithm on datasets with a single face in each image, therefore, this correlation property is a feature. We have shown that this technique
works very well for images with single face and controlled backgrounds but not for complex backgrounds and uncontrolled features involving pose, location and lighting. We do not present this technique for face detection in images taken under uncontrolled conditions and do not claim its strength in such data sets. There are techniques, for example Viola and Jones face detector, that
are better suited for such datasets.

However, the strength of our correlation based face detector lies in specific and repeated settings. We have clearly demonstrated that, with controlled location and small variation in scale, pose and lighting, our technique significantly outperforms Viola and Jones' face detector. 

Prior to this research, there has been a lot of study on correlation filters and their applications but this is the first research that extends correlation filters to face detection. The goal is not to create the best face detection approach
for all scenarios, but to extend correlation filters to an area where they have never been applied before. More specifically, this face detector has been designed to work under very controlled scenarios, specific to a situation. It has a potential value in scenarios such as, an airport security gate, an entrance to a sensitive building such as a
nuclear reactor or an entrance to a train / bus station. We end this paper by summarizing the main findings of this research:
\begin{enumerate}
\item It is a fast algorithm suitable for face detection in cameras.
\item This is the first correlation based face detector. We have presented scenarios where such a face detector would
be successful and where it may not be an approach of choice.
\item We have presented a methodology to use localization for face detection.
\item A correlation filter trained for a given scale and pose can be used for a small range of scales and poses around
the training scale and pose.
\item Correlation based face detectors are most promising in deployments where the setting remains roughly constant and setting specific filters may learn to discount background distractions.
\end{enumerate}

\ifCLASSOPTIONcaptionsoff
  \newpage
\fi



%


{\small
\bibliographystyle{ieee}
\bibliography{facerec}

\begin{thebibliography}{10}\itemsep=-1pt

\bibitem{pasc}
J.~R. Beveridge, P.~J. Phillips, D.~Bolme, B.~A. Draper, G.~H. Givens, Y.~M.
  Lui, M.~N. Teli, H.~Zhang, T.~Scruggs, S.~Cheng, K.~Bowyer, and P.~Flynn.
\newblock {The Point-and- Shoot Challenge}.
\newblock In {\em Sixth IEEE International Conference on Biometrics: Theory
  Applications and Systems (BTAS)}, September/October 2013.

\bibitem{bolmecvpr10}
D.~Bolme, J.~Beveridge, B.~Draper, and Y.~Lui.
\newblock Visual object tracking using adaptive correlation filters.
\newblock In {\em CVPR}, 2010.

\bibitem{asef}
D.~S. Bolme, B.~A. Draper, and J.~R. Beveridge.
\newblock Average of synthetic exact filters.
\newblock In {\em {IEEE Conference on Computer Vision and Pattern Recognition
  (CVPR)}}, pages 2105--2112, August 2009.

\bibitem{opencv_library}
G.~Bradski.
\newblock {The OpenCV Library}.
\newblock {\em Dr. Dobb's Journal of Software Tools}, 2000.

\bibitem{caulfield}
H.~J. Caulfield.
\newblock Role of the horner efficiency in the optimization of spatial filters
  for optical pattern recognition.
\newblock {\em Applied Optics}, 21:4391--4392, 1982.

\bibitem{chencvpr2014}
S.~Chen, D.and~Ren, Y.~Wei, X.~Cao, and J.~Sun.
\newblock Joint cascade face detection and alignment.
\newblock In {\em CVPR}, 2014.

\bibitem{cheney15unconstrained}
J.~Cheney, B.~Klein, A.~K. Jain, and B.~F. Klare.
\newblock Unconstrained face detection: State of the art baseline and
  challenges.
\newblock In {\em 8th International Conference on Biometrics}, May 2015.

\bibitem{crow1984}
F.~C. Crow.
\newblock {Summed-Area Tables for Texture Mapping }.
\newblock {\em ACM SIGGRAPH Computer Graphics}, 18(3):207--212, 1984.

\bibitem{Farfade:2015}
S.~S. Farfade, M.~J. Saberian, and L.-J. Li.
\newblock Multi-view face detection using deep convolutional neural networks.
\newblock In {\em Proceedings of the 5th ACM on International Conference on
  Multimedia Retrieval}, ICMR '15, pages 643--650, New York, NY, USA, 2015.
  ACM.

\bibitem{felzenszwalbcvpr10}
P.~Felzenszwalb, R.~Girshick, and D.~McAllester.
\newblock Cascade object detection with deformable part models.
\newblock In {\em CVPR}, 2010.

\bibitem{forsyth}
D.~Forsyth and J.~Ponce.
\newblock {\em Computer Vision: A Modern Approach}.
\newblock Prentice-Hall, 2002.

\bibitem{freund}
Y.~Freund and R.~E. Schapire.
\newblock A decision-theoretic generalization of on-line learning and an
  application to boosting.
\newblock In {\em Computational Learning Theory: Eurocolt 95}, pages 23--37.
  Springer-Verlag, 1995.

\bibitem{goo96}
G.~{Goo}.
\newblock Correlation filter for target detection and noise and clutter
  rejection.
\newblock In D.~P.~C. . T.-H. Chao, editor, {\em Society of Photo-Optical
  Instrumentation Engineers (SPIE) Conference Series}, volume 2752 of {\em
  Society of Photo-Optical Instrumentation Engineers (SPIE) Conference Series},
  pages 317--332, March 1996.

\bibitem{sdf}
C.~F. Hester and D.~Casasent.
\newblock Multivariant technique for multiclass pattern recognition.
\newblock {\em Applied Optics}, 19(11):1758--1761, 1980.

\bibitem{horner}
J.~L. Horner.
\newblock Light utilization in optical correlators.
\newblock {\em Applied Optics}, 21:4511--4514, 1982.

\bibitem{ijcbpasc2014}
{J. Ross Beveridge and Hao Zhang and Patrick J. Flynn and Yooyoung Lee and
  Venice Erin Liong and Jiwen Lu and Marcus De Assis Angeloni and Tiago de
  Freitas Pereira and Haoxiang Li and Gang Hua and Vitomir Struc and Janez
  Krizaj and P. Jonathon Phillips}.
\newblock {The IJCB 2014 PaSC Video Face and Person Recognition Competition}.
\newblock In {\em International Joint Conference on Biometrics}, 2014.

\bibitem{klare2015pushing}
B.~F. Klare, B.~Klein, E.~Taborsky, A.~Blanton, J.~Cheney, K.~Allen,
  P.~Grother, A.~Mah, M.~Burge, and A.~K. Jain.
\newblock Pushing the frontiers of unconstrained face detection and
  recognition: Iarpa janus benchmark a.
\newblock In {\em 2015 IEEE Conference on Computer Vision and Pattern
  Recognition (CVPR)}, pages 1931--1939. IEEE, 2015.

\bibitem{mvsdf}
B.~V. Kumar.
\newblock Minimum-variance synthetic discriminant functions.
\newblock {\em Journal of Optical Society of America}, 3(10):1579--1584,
  October 1986.

\bibitem{corrFil}
B.~V. Kumar.
\newblock Tutorial survey of composite filter designs for optical correlators.
\newblock {\em Appl. Opt.}, 31:4773--4801, 1992.

\bibitem{kumar2007}
B.~V. K.~V. Kumar, M.~Savvides, and C.~Xie.
\newblock {Correlation Pattern Recognition for Face Recognition}.
\newblock {\em Proceedings of the IEEE}, 94(11):1963--1976, 2007.

\bibitem{deeplearningnature}
Y.~LeCun, Y.~Bengio, and G.~Hinton.
\newblock Deep learning.
\newblock {\em Nature}, 521(7553):436--444, 05 2015.

\bibitem{Lincvpr2014}
G.~Li, H.and~Hua, J.~Brandt, and J.~Yang.
\newblock Efficient boosted exemplar-based face detection.
\newblock In {\em CVPR}, 2014.

\bibitem{Licvpr2015}
Z.~Li, H.and~Lin, X.~Shen, J.~Brandt, and G.~Hua.
\newblock A convolutional neural network cascade for face detection.
\newblock In {\em CVPR}, 2015.

\bibitem{savvides}
B.~V.~K. M.~Savvides and P.~Khosla.
\newblock Face verification using correlation filters.
\newblock In {\em Third IEEE Automatic Identification Advanced Technologies},
  pages 56--61, Tarrytown, NY, 2002.

\bibitem{umace}
A.~Mahalanobis, B.~V.~K. Kumar, S.~Song, S.~R.~F. Sims, and J.~F. Epperson.
\newblock Unconstrained correlation filters.
\newblock {\em Applied Optics}, 33(17):3751--3759, June 1994.

\bibitem{polyfilter}
A.~Mahalanobis and B.~V. K.~V. Kumar.
\newblock {Polynomial Filters for Higher Order Correlation and Multi-input
  Information Fusion}.
\newblock In {\em Proc. SPIE, Euro-American Optoelectronic Information
  Processing Workshop}, pages 221--231, 1997.

\bibitem{mahalanobis}
A.~Mahalanobis, B.~V. K.~V. Kumar, and D.~Casasent.
\newblock Minimum average correlation energy filters.
\newblock {\em APPLIED OPTICS}, 26(17):3633--3640, 1987.

\bibitem{dccf}
A.~Mahalanobis, B.~V. K.~V. Kumar, and S.~R.~F. Sims.
\newblock {Distance Classifier Correlation Filters For Multi-class Target
  Recognition}.
\newblock {\em Applied Optics}, 35:3127--3133, 1996.

\bibitem{north63}
D.~O. North.
\newblock An {A}nalysis of the {F}actors which {D}etermine {S}ignal/{N}oise
  {D}iscriminations in {P}ulsed {C}arrier {S}ystems.
\newblock In {\em Proc. IEEE}, volume~51, pages 1016--1027, July 1963.

\bibitem{pascChallenge2016}
{Patrick J. Flynn, P. Jonathon Phillips, and Walter J. Scheirer}.
\newblock {Video Person Recognition Evaluation}.
\newblock In {\em 8th IEEE International Conference on Biometrics: Theory,
  Applications, and Systems}, 2016.

\bibitem{feret}
P.~J. Phillips, H.~Moon, S.~A. Rizvi, and P.~J. Rauss.
\newblock The feret evaluation methodology for face-recognition algorithms.
\newblock {\em IEEE Transaction on Pattern Analysis and Machine Intelligence
  (PAMI)}, 22(10):1090--1104, 2000.

\bibitem{burtAdelson}
{P.J. Burt and E. H. Adelson}.
\newblock {The laplacian pyramid as a compact image code}.
\newblock {\em IEEE Transaction Communications}, 31:532--549, 1983.

\bibitem{refregier}
P.~Refregier.
\newblock Optimal trade-off filters for noise robustness, sharpness of the
  correlation peak, and horner efficiency.
\newblock {\em OPTICS LETTERS}, 16(11):829--831, June 1991.

\bibitem{kerekes}
R.Kerekes and B.~V. K.~V. Kumar.
\newblock {Correlation Filters with Controlled Scale Response}.
\newblock {\em IEEE Trans. Image Process}, 15, July 2006.

\bibitem{mellin}
J.~Rosen and J.~Shamir.
\newblock {Scale-Invariant Pattern Recognition with Logarithmic Radial Harmonic
  Filters}.
\newblock {\em Applied Optics}, 28:240--244, 1989.

\bibitem{fgpasc2015}
{Ross Beveridge}, {Hao Zhang}, {Bruce Draper}, {Patrick Flynn}, {Zhenhua Feng},
  {Patrik Huber}, J.~Kittler, Z.~Huang, S.~Li, Y.~Li, M.~Kan, R.~Wang, S.~Shan,
  X.~Chen, H.~Li, G.~Hua, V.~Struc, J.~Krizaj, C.~Ding, D.~Tao, and
  J.~Phillips.
\newblock {Report on the FG 2015 Video Person Recognition Evaluation}.
\newblock In {\em {The Eleventh IEEE International Conference on Automatic Face
  and Gesture Recognition (FG)}}, 2015.

\bibitem{mace}
M.~Savvides, B.~V. Kumar, and P.~Khosla.
\newblock Face verification using correlation filters.
\newblock In {\em Proc. Of Third IEEE Automatic Identification Advanced
  Technologies}, pages 56--61, Tarrytown, NY, 2002.

\bibitem{savvides03}
M.~Savvides, K.~Venkataramani, and B.~Vijaya~Kumar.
\newblock Incremental updating of advanced correlation filters for biometric
  authentication systems.
\newblock In {\em IEEE International Conference on Multimedia and Expo},
  volume~3, pages 229--232, 2003.

\bibitem{savvidesKumar}
M.~Savvides and B.~Vijaya~Kumar.
\newblock Efficient design of advanced correlation filters for robust
  distortion-tolerant face recognition.
\newblock In {\em IEEE Conference on Advanced Video and Signal Based
  Surveillance}, pages 45--52, 2003.

\bibitem{kumar02}
B.~VijayaKumar, M.~Savvides, K.~Venkataramani, and C.~Xie.
\newblock Spatial frequency domain image processing for biometric recognition.
\newblock In {\em IEEE International Conference on Image Processing}, volume~1,
  pages 53--56, 2002.

\bibitem{viola-jones}
P.~Viola and M.~Jones.
\newblock Rapid object detection using a boosted cascade of features.
\newblock In {\em Computer Vision and Pattern Recognition (CVPR)}, 2001.

\bibitem{viola-jonesJournal}
P.~Viola and M.~J. Jones.
\newblock Robust real-time face detection.
\newblock {\em International Journal of Computer Vision}, 57(2):137--154, 2004.

\bibitem{yangkriegmanahuja}
M.-H. Yang, D.~Kriegman, and N.~Ahuja.
\newblock Detecting faces in images: A survey.
\newblock {\em IEEE Transaction on PAMI}, 24(1):34--58, 2002.

\bibitem{yangiccv2015}
P.~Yang, S.and~Luo, C.~Loy, and X.~Tang.
\newblock From facial parts responses to face detection: A deep learning
  approach.
\newblock In {\em ICCV}, 2015.

\bibitem{zhang2016}
K.~Zhang, Z.~Zhang, and Z.~Li.
\newblock Joint face detection and alignment using multi-task cascaded
  convolutional networks.
\newblock In {\em ArXiv Report}, 2016.

\end{thebibliography}


\begin{thebibliography}{1}\itemsep=-1pt

\bibitem{forsyth}
D.~Forsyth and J.~Ponce.
\newblock {\em Computer Vision: A Modern Approach}.
\newblock Prentice-Hall, 2002.

\bibitem{viola-jones}
P.~Viola and M.~Jones.
\newblock Rapid object detection using a boosted cascade of features.
\newblock In {\em Computer Vision and Pattern Recognition (CVPR)}, 2001.

\end{thebibliography}
}
%

\begin{IEEEbiography}[{\includegraphics[width=1in,height=1.20in,clip,keepaspectratio]{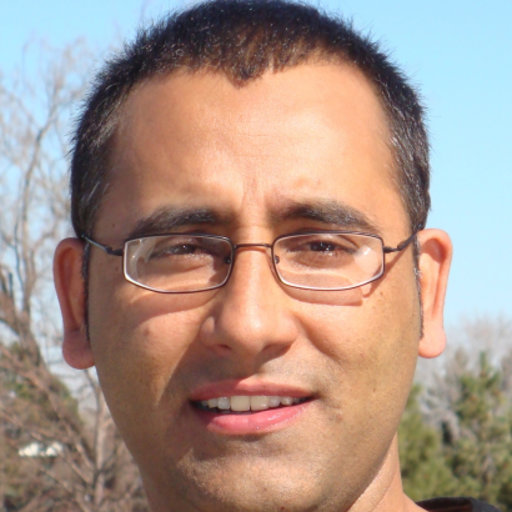}}]{Mohammad Nayeem Teli}
Dr. Teli's research interests center on improving the understanding and design of systems employing the use of Computer Vision, Data science, Machine Learning and Mathematical Modeling algorithms. He has worked on design and development of face recognition algorithms to help understand hard datasets. Dr. Teli is a lecturer teaching Computer Science at the University of Maryland, College Park. He is also a co-founder of a startup involved in Data Science Research and Development.
\end{IEEEbiography}
\begin{IEEEbiography}[{\includegraphics[width=1in,height=1.20in,clip,keepaspectratio]
{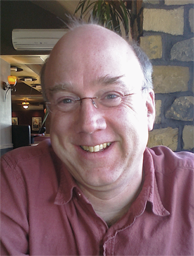}}]{Bruce A. Draper}
 Dr. Draper is a professor of Computer Science at Colorado State University; Professor Draper also holds an appointment in the Molecular, Cellular and Integrated Neuroscience (MCIN) program. Professor Draper's research interests are in biologically inspired computer vision, action recognition, and face recognition, and co-directs the Computer Vision Laboratory with Professor Ross Beveridge. He also teaches at both the graduate and undergraduate levels.
\end{IEEEbiography}
\begin{IEEEbiography}[{\includegraphics[width=1in,height=1.20in,clip,keepaspectratio]{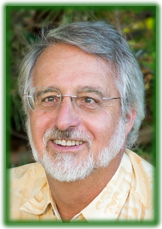}}]{J. Ross Beveridge}
Professor Beveridge works on Computer Vision, emphasizing problems relating to recognition and understanding. His current work draws upon past work in video understanding, human face recognition and object recognition. Evaluation is one key part of this past work, including face recognition in video, and older legacy work supplying open source tools such as the CSU Face Identification Evaluation System. Professor Beveridge's other interests include high dimensional data analysis, optimal matching of geometric features, genetic algorithms and the use of reconfigurable embedded hardware.
\end{IEEEbiography}
\vfill



\end{document}